\documentclass{article}

\usepackage[preprint]{neurips_2026}

\usepackage[utf8]{inputenc}
\usepackage[T1]{fontenc}
\PassOptionsToPackage{hyphens}{url}
\usepackage[hidelinks,pageanchor=false,breaklinks=true,bookmarks=false]{hyperref}
\usepackage{url}
\usepackage{booktabs}
\usepackage{amsfonts}
\usepackage{amsmath}
\usepackage{amssymb}
\usepackage{nicefrac}
\usepackage{microtype}
\usepackage{xcolor}
\usepackage{graphicx}
\usepackage{multirow}
\usepackage{makecell}
\usepackage{bm}
\usepackage{float}
\graphicspath{{figures/}{./}}

\title{Cross-Model KV Cache Transfer in LLM Families:\\A Closed-Form Linear Mapping for Prefill Reuse}

\author{%
  Taekyung Heo \quad Rasoul Shafipour \quad Ritchie Zhao \quad Maximilian Golub \\
  \bf Mohammad Mahdi Kamani \quad Ritika Borkar \quad Makesh Tarun Chandran \\
  \bf Pantea Zardoshti \quad Bita Darvish Rouhani \\[3pt]
  \normalfont NVIDIA
}

\begin{document}

\maketitle

\begin{abstract}
    Production deployments often swap between different-sized models in a family for cost-quality cascading, mid-conversation switching, and routing, and each swap forces the receiver to repay the prefill from scratch. We propose \emph{cross-model KV cache transfer}, where the receiver reuses the source's KV cache, skipping prefill. We find that cross-model KV has substantial linear structure across matched-KV pairs, where source and target share KV head count and per-head dimension. On Qwen3 14B$\to$32B, one source layer explains 56\% of variance in the target's keys and 32\% in values, rising to 79\% and 65\% with multiple source layers. Building on this, we design a closed-form ridge mapper that operates per head and proceeds in three steps. First, for each target layer we select the top-$k$ most predictive source layers and concatenate their KV as input. Second, we strip RoPE from the keys before mapping, so the fit is position-free and reusable across context lengths. Third, we fit ridge regression on a small calibration set of 500 FineWeb-Edu sequences of 1{,}024 tokens each. Surprisingly, across six pairs in three families, this linear mapper retains 73--98\% of the receiver's standalone-prefill accuracy on four pairs, while two degrade sharply. A nonlinear MLP recovers up to $+37$\,pp HellaSwag retention on the failures. The mapper runs 2.7--25$\times$ faster than re-prefill and remains stable across multi-turn handoff, making cross-model KV cache transfer practical.
\end{abstract}

\section{Introduction}
\label{sec:intro}

Production LLM serving increasingly involves long agentic sessions, where context accumulates across many turns. It also relies on multi-model orchestration for cost-quality cascading, mid-conversation switching, and routing \citep{openai2025gpt5}. These practices swap between different-sized members of a \emph{model family} \citep{yang2025qwen3,grattafiori2024llama}. Family members share core architectural choices and substantial pretraining data, but differ in architectural details and training recipe across scales. Both trends compound \emph{prefill} cost. Long sessions stretch the prompt, and each model swap re-prefills the accumulated context on the receiver. Prefill is the forward pass that populates the KV cache before generation. Its cost scales with model size and prompt length. Prefix caching mitigates this within a single model only.

\begin{figure}[t]
  \centering
  \includegraphics[width=0.95\linewidth]{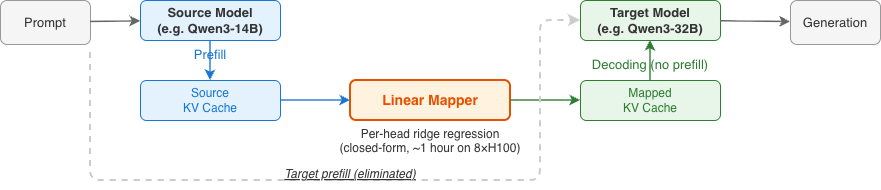}
  \vspace{-0.4em}
  \caption{\textbf{Cross-model KV cache transfer pipeline.} A per-head linear map converts the source's prefilled KV cache into the target's expected format, so the target decodes without re-prefilling.}
  \label{fig:pipeline}
  \vspace{-0.3em}
\end{figure}

Since prefill's output is the KV cache, reusing it across models reduces to a
representation problem of transforming one model's KV cache into another's expected
format. We call this \emph{cross-model KV cache transfer} and restrict the present study
to within-family transfer. The mapping is bidirectional. Small-to-large transfer
upgrades quality, and large-to-small reduces cost. Figure~\ref{fig:pipeline} illustrates the pipeline.
This pipeline generalizes intra-model KV reuse. Cross-layer sharing
\citep{brandon2024cla,wu2024lckv,chang2025xkv} exploits redundancy within a
model, and prefix caching \citep{zheng2024sglang,gim2024promptcache} reuses KV across
requests of the same model. Cross-model
KV cache transfer maps the cache values \emph{from one model to another}.

Cross-model KV cache transfer is challenging because source and target can differ in
layer count, hidden dimension, and KV head configuration. Prior work bridges these
gaps with neural fusers \citep{fu2026c2c}, learned latent-space adapters
\citep{dery2026latentalign}, attention-pattern mapping \citep{zhao2025iam}, or
same-architecture KV sharing \citep{liu2026droidspeak}, each requiring
gradient-based training or strong architectural assumptions (\S\ref{sec:prior}).

In this study, we show that the cross-model KV relationship exhibits substantial linear structure, admitting a small-sample, gradient-free fit. On Qwen3 14B$\to$32B, a single source layer explains 56\% of variance in the target's keys and 32\% in values, rising to 79\% and 65\% with multiple source layers (\S\ref{sec:linear_structure}). Building on this observation, we propose a closed-form per-head ridge mapper (\S\ref{sec:design}) that combines three components: per-head ridge regression fit from a small calibration set, cross-layer source selection where each target layer draws from its top-$k$ most predictive source layers, and content-space (RoPE-stripped) mapping that decouples positional rotation from semantic content so the fit transfers across context lengths.

Across six \emph{matched-KV} pairs from three families, where source and target share KV head count and per-head dimension (\S\ref{sec:main}), the linear mapper retains 73--98\% of standalone accuracy averaged across five benchmarks on its best pairs. We further find that error placement, not error magnitude, determines per-pair retention. A nonlinear MLP redistributes residual error away from attention-sensitive subspaces and adds up to $+37$\,pp HellaSwag retention on harder pairs (\S\ref{sec:nonlin}). Across 12 matched-KV pair evaluations from three families, attention-output similarity correlates with HellaSwag retention at Pearson $r{=}{+}0.57$ and outperforms $R^2$ (\S\ref{sec:mechanism}).

\vspace{-0.3em}
\paragraph{Contributions.}
(1)~\textbf{Closed-form mapping framework.} We propose a gradient-free framework
for cross-model KV cache transfer based on per-head ridge regression. Two design
choices are central: each target layer is fit from its top-$k$ most predictive
source layers, and keys are mapped in RoPE-stripped content space so the fit
transfers across context lengths (\S\ref{sec:design}).
(2)~\textbf{Multi-family validation.} We validate the framework on six
matched-KV pairs across three families. Four pairs retain 73--98\% of
standalone accuracy on five benchmarks at 2.7--25$\times$ lower prefill
latency than re-prefill in the small-to-large direction (\S\ref{sec:latency}).
(3)~\textbf{Nonlinear extension.} A nonlinear MLP adds up to
$+37$\,pp HellaSwag retention on harder Ministral pairs by redistributing
residual error away from attention-sensitive subspaces
(\S\ref{sec:nonlin}, \S\ref{sec:mechanism}).
(4)~\textbf{Attention-output cosine predicts cross-pair retention.} Across 12
matched-KV pair evaluations from three families, attention-output cosine
correlates with HellaSwag retention at Pearson $r{=}{+}0.57$, outperforming
$R^2$ ($r{=}{-}0.20$). The same concentration analysis explains
the $+37$\,pp MLP gain as error redistribution toward attention-irrelevant
directions (\S\ref{sec:mechanism}).

\section{Background and motivation}
\label{sec:background}

\subsection{Problem formulation}
\label{sec:formulation}

Consider a source model $\mathcal{S}$ with $L_s$ layers and a target model $\mathcal{T}$
with $L_t$ layers. Both use grouped-query attention with $n_{\text{kv}}^s$ and
$n_{\text{kv}}^t$ KV heads of dimension $d_h^s$ and $d_h^t$, respectively. For an input
sequence $\mathbf{x}=(x_1,\ldots,x_T)$ of length $T$, layer~$l\in\{1,\ldots,L_s\}$,
head~$h\in\{1,\ldots,n_{\text{kv}}^s\}$ of the source model produces keys and values
$\mathbf{K}_s^{l,h},\mathbf{V}_s^{l,h}\in\mathbb{R}^{T\times d_h^s}$. We write
$\mathcal{C}_{\mathcal{S}}$ for the full source KV cache across all layers and
heads, and $\mathcal{C}_{\mathcal{T}}$ analogously for the target. Source and
target share a tokenizer within a family, so the input sequence has the same
length $T$ for both. We say a transfer pair has \emph{matched KV} when
$n_{\text{kv}}^s = n_{\text{kv}}^t$ and $d_h^s = d_h^t$, even if $L_s$, $L_t$,
or parameter counts differ across scales.
\vspace{-3pt}

We seek a mapping $f\colon \mathcal{C}_{\mathcal{S}}\to\hat{\mathcal{C}}_{\mathcal{T}}$
such that the target model, decoding from $\hat{\mathcal{C}}_{\mathcal{T}}$ in place of
its own $\mathcal{C}_{\mathcal{T}}$, produces equivalent outputs:
\begin{equation}
m\bigl(\mathcal{T}(\mathbf{x};\hat{\mathcal{C}}_{\mathcal{T}})\bigr)
\;\approx\;
m\bigl(\mathcal{T}(\mathbf{x};\mathcal{C}_{\mathcal{T}})\bigr),
\end{equation}
for a downstream task $\tau$ with metric~$m$. We decompose $f$ into per-head mappings
$f_K^{l,h}$ and $f_V^{l,h}$ that produce the target's keys and values for each layer and
head. We measure transfer quality by the target model's downstream accuracy, not by
reconstruction metrics alone.

\vspace{-4pt}
\subsection{Prior approaches}
\label{sec:prior}
Recent cross-model KV-reuse methods each require either gradient-based training or
architectural constraints. C2C \citep{fu2026c2c} trains per-pair neural fusers.
LatentAlign \citep{dery2026latentalign} learns per-model adapters into a shared
latent space. IAM \citep{zhao2025iam} substitutes small-model attention
\emph{patterns}, not KV values. DroidSpeak \citep{liu2026droidspeak} requires
identical architecture. Further training-time approaches
\citep{woo2026prefillshare,woo2026icarus} share the same gradient-based
constraint. To the best of our knowledge, none investigates whether the
cross-model KV relationship is simple enough for a closed-form, training-free
mapping. Three orthogonal lines compose with the present work: intra-model
cross-layer KV reuse \citep{brandon2024cla,wu2024lckv,chang2025xkv}, prefill
acceleration \citep{liu2025specprefill,upasani2026xfsp,qiao2025swiftkv}, and
linear cross-LLM representation alignment
\citep{huh2024platonic,chen2025stitching,bello2025lrt,huang2025crossmodel}.
Table~\ref{tab:prior} (Appendix~\ref{app:prior_comparison}) summarizes the
cross-model comparison.

\vspace{-4pt}
\subsection{Linear structure in cross-model KV}
\label{sec:linear_structure}

\begin{figure}[t]
  \centering
  \IfFileExists{figures/r2_heatmaps.png}
    {\includegraphics[width=\linewidth]{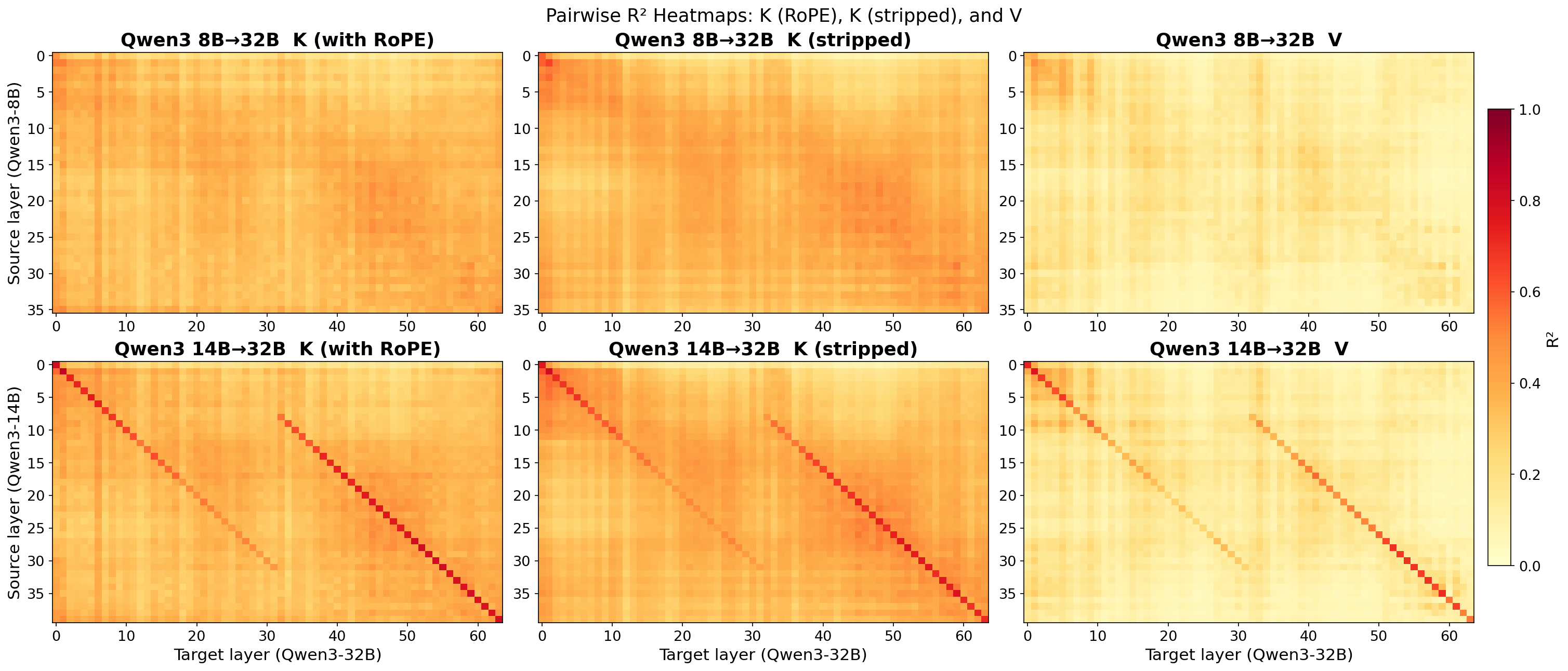}}
    {\fbox{\rule{\linewidth}{0pt}\rule{0pt}{4cm}}}
    \vspace{-0.8em}
  \caption{\textbf{Cross-model KV exhibits substantial linear structure.}
  Each heatmap cell shows head-averaged $R^2$ from a single-source linear
  regression mapping source layer $l'$ (row) to target layer $l$ (column),
  for raw keys ($K_{\text{rope}}$), RoPE-stripped keys ($K_{\text{stripped}}$),
  and values ($V$).}
  \label{fig:r2-heatmaps}
\end{figure}

Before committing to a mapper architecture, we ask what structure the cross-model
KV relationship has. We analyze this structure on Qwen3. The
linear-structure result generalizes to other matched-KV pairs. We probe three cache types. $K_{\text{rope}}$ denotes keys as the model uses
them. $K_{\text{stripped}}$ denotes keys with the position-dependent RoPE
rotation removed via $\mathbf{R}_{\Theta}^{-1}$ (see \S\ref{sec:rope}). $V$
denotes values, which carry no positional encoding. For each cache type
$C\in\{K_{\text{rope}},K_{\text{stripped}},V\}$ and each triple of source layer
$l'\in\{1,\ldots,L_s\}$, target layer $l\in\{1,\ldots,L_t\}$, and head $h$, we
fit a single-source ordinary least-squares regression at the token level. Each
of the $N$ calibration tokens, subsampled from FineWeb-Edu sequences as detailed
in \S\ref{sec:ridge}, contributes one observation, mapping the source's
per-token feature vector $C_s^{l',h}\in\mathbb{R}^{d_h^s}$ to the target's
$C_t^{l,h}\in\mathbb{R}^{d_h^t}$,
\begin{equation}
  \hat{C}_t^{l,h} = C_s^{l',h}\,\mathbf{W} + \mathbf{b},
  \qquad
  \mathbf{W}\in\mathbb{R}^{d_h^s\times d_h^t},\ \mathbf{b}\in\mathbb{R}^{d_h^t},
  \label{eq:single_source_ols}
\end{equation}
treating $C_s^{l',h}$ as a row vector to match the production-mapper convention in
\S\ref{sec:design}.
This is a per-(triple, cache type) probe, not the production mapper of
\S\ref{sec:design}, which concatenates several source layers and adds ridge
regularization. We measure fit quality by the coefficient of determination $R^2$,
averaged across the $n_{\text{kv}}^t$ target heads to give a head-averaged $R^2$
for each $(l',l)$ pair. Visualizing this as a heatmap with rows indexing source
layers and columns indexing target layers (Figure~\ref{fig:r2-heatmaps}) reveals
four qualitative patterns. Per-pair quantitative numbers are reported in
\S\ref{sec:experiments}.

\textbf{(1)~High linear fit:} a single
linear regression already captures a large fraction of cross-model KV
variance on the most predictable target layers. Head-averaged $R^2$ is
well above zero along the diagonal, and the best single source--target
cell reaches $K_{\text{stripped}}$ $R^2{=}0.81$ on Qwen3 14B$\to$32B
and $0.65$ on the weaker 8B$\to$32B at the heatmap peaks. Layer-averaged
production-ridge $R^2$ is reported in Appendix~\ref{app:ablation}.
\textbf{(2)~Closer models give sharper diagonals:} larger architectural and depth
gaps diffuse the pattern. \textbf{(3)~RoPE contaminates the fit:} stripping RoPE
generally sharpens the diagonal, motivating the position--content decomposition in
\S\ref{sec:rope}. \textbf{(4)~K is more predictable than V:} typically a $\sim$0.2
gap in head-averaged $R^2$ (Appendix~\ref{app:linear_structure}).

\paragraph{How many source layers does the mapper need?}
The probe uses a \emph{single} source layer. We extend it with greedy
forward selection, iteratively adding the source layer that maximally
increases $R^2$. Complementary information is distributed across multiple
source layers. On Qwen3 14B$\to$32B, $k{=}1$ captures only $66\%$ of the
$k{=}\text{all}$ $R^2$ for $K_{\text{stripped}}$ and $42\%$ for $V$, with
the largest gain from $k{=}1$ to $k{=}4$ and $R^2$ close to its
$k{=}\text{all}$ value by $k{=}6$
(Appendices~\ref{app:linear_structure} and~\ref{app:ablation}).
This motivates the top-$k$ selection (\S\ref{sec:topk}).

\section{Mapper design}
\label{sec:design}

\begin{figure}[t]
  \centering
  \IfFileExists{figures/mapper_architecture.drawio.png}
    {\includegraphics[width=\linewidth]{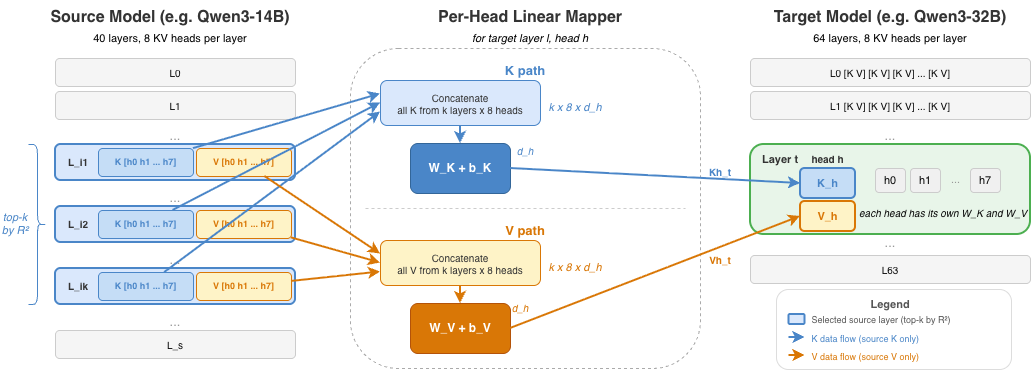}}
    {\fbox{\rule{\linewidth}{0pt}\rule{0pt}{3cm}}}
  \vspace{-0.8em}
  \caption{\textbf{Per-head linear mapper.} For each target $(l, h)$, the
  top-$k$ source layers (selected per target layer by head-averaged $R^2$)
  are concatenated. Independent ridge regressions $\mathbf{W}_K^{l,h}$ and
  $\mathbf{W}_V^{l,h}$ project to the target's K and V spaces. No
  parameters are shared across heads or between K and V.}
  \label{fig:mapper}
\end{figure}

Figure~\ref{fig:mapper} illustrates the mapper for one target (layer~$l$, head~$h$)
pair. A top-$k$ subset of source layers is selected per target layer by $R^2$, the
selected source keys and values are concatenated, and independent ridge regressions
$\mathbf{W}_K$, $\mathbf{W}_V$ map them to the target. Each mapper is a closed-form
linear solve repeated independently for all $L_t\times n_{\text{kv}}^t$ target heads.
The mapper has three components: per-head ridge (\S\ref{sec:ridge}), cross-layer source
selection (\S\ref{sec:topk}), and content-space mapping (\S\ref{sec:rope}).

\vspace{-4pt}
\subsection{Per-head ridge regression}
\label{sec:ridge}

Source and target differ in layer count ($L_s\ne L_t$), head dimension
($d_h^s\ne d_h^t$), and potentially KV head count
($n_{\text{kv}}^s\ne n_{\text{kv}}^t$). We address this by fitting an independent linear
mapping per target (layer, head):
\begin{equation}
  \hat{\mathbf{K}}_t^{l,h} = \mathbf{X}_K^{l}\mathbf{W}_K^{l,h} + \mathbf{b}_K^{l,h},
  \qquad
  \hat{\mathbf{V}}_t^{l,h} = \mathbf{X}_V^{l}\mathbf{W}_V^{l,h} + \mathbf{b}_V^{l,h},
  \label{eq:ridge_kv}
\end{equation}
where $\mathbf{X}_K^{l}$ concatenates source key features from the $k$ selected
source layers (\S\ref{sec:topk}) and $\mathbf{W}_K^{l,h}\in\mathbb{R}^{(k\cdot
n_{\text{kv}}^s\cdot d_h^s)\times d_h^t}$ is the weight matrix.

\vspace{-3pt}

\paragraph{Fitting.} Stacking $N$ calibration tokens gives a design matrix
$\mathbf{X}\in\mathbb{R}^{N\times(k\, n_{\text{kv}}^s\, d_h^s)}$ and a response
matrix $\mathbf{Y}\in\mathbb{R}^{N\times d_h^t}$. We use ridge with $\lambda=0.01$
rather than pure OLS for numerical conditioning. The feature dimension can reach
tens of thousands at large~$k$ and the top-$k$ selected source layers are by
construction correlated, so $\mathbf{X}^\top\mathbf{X}$ is near-singular. The
Tikhonov term stabilizes the inverse with negligible fit bias. The closed-form
solution is
\begin{equation}
  \mathbf{W}^* = (\mathbf{X}^\top\mathbf{X}+\lambda\mathbf{I})^{-1}\mathbf{X}^\top\mathbf{Y}.
  \label{eq:Wstar}
\end{equation}
We center $\mathbf{X}$ and $\mathbf{Y}$ before solving, so $\mathbf{W}^*$ above
estimates the slope and the bias in Eq.~\ref{eq:ridge_kv} is recovered as
$\mathbf{b} = \bar{\mathbf{Y}} - \bar{\mathbf{X}}\mathbf{W}^*$.
Calibration data consists of 500 FineWeb-Edu sequences of length 1024, stride-4 subsampled to
$N\approx 128$K tokens per target head. Total mapper size (K plus V) is
$2\,L_t\,n_{\text{kv}}^t\,(k\, n_{\text{kv}}^s\, d_h^s)\,d_h^t$, with per-pair
sizes in Appendix~\ref{app:setup}. Fitting takes $\sim$47--87\,min per pair
on a single $8\times$H100 node with no gradient-based training. Fitting-time
scaling is sub-linear in the target head count because forming $\mathbf{X}^\top\mathbf{X}$
($\mathcal{O}(N d_s^2)$, with $d_s = k\,n_{\text{kv}}^s\,d_h^s$ the source feature dimension) dominates wall time and is computed once per target
layer, shared across its heads. Per-pair times are in Appendix~\ref{app:run_info}.

\vspace{-4pt}
\subsection{Cross-layer source selection}
\label{sec:topk}

The source and target have different numbers of layers, raising the question of which
source layers should inform each target layer's mapping. \S\ref{sec:linear_structure}
showed that not all source layers are equally informative for a given target layer. For
each target layer~$l$, we therefore select the top-$k$ source layers by
head-averaged $R^2$ averaged over RoPE-stripped keys and values, concatenate their KV
features, and fit the multi-source ridge regression of \S\ref{sec:ridge}:
\begin{equation}
  \mathbf{X}_K^{l} = [\bar{\mathbf{K}}_s^{l_1}\,\|\,\bar{\mathbf{K}}_s^{l_2}\,\|\,\cdots\,\|\,\bar{\mathbf{K}}_s^{l_k}],
\end{equation}
where $\bar{\mathbf{K}}_s^{l_i}\in\mathbb{R}^{T\times(n_{\text{kv}}^s\cdot d_h^s)}$ is
the concatenation of all Key heads from source layer~$l_i$, and $\{l_1,\ldots,l_k\}$ are
the top-$k$ source layers for target layer~$l$. $\mathbf{X}_V^l$ is constructed
analogously from Value heads. All heads within a target layer share
the same selected source layers, enabling cross-head information flow. The number of
source layers $k$ is a hyperparameter selected per pair by a sweep
(\S\ref{sec:setup}). Cross-layer source selection is the largest single contributor
among the three mapper components (\S\ref{sec:ablation}).

\vspace{-4pt}
\subsection{Content-space mapping (RoPE factoring)}
\label{sec:rope}

Standard RoPE \citep{su2024rope} applies a position-dependent rotation to
queries and keys. The KV cache stores only the rotated keys
$\mathbf{k}_{\text{RoPE}}(t)=\mathbf{R}_{\Theta}(t)\,\mathbf{k}_{\text{content}}$,
while queries are recomputed at decode time.
Decoupling RoPE from content makes the mapping pipeline modular. The per-head
ridge weights are computed in a position-free space, so they remain valid
across pairs with different RoPE configurations and at any position the
target's RoPE supports. Fitting the mapper directly on RoPE-coupled keys also
works within noise on short-context benchmarks (Table~\ref{tab:ablation}), but
ties the weights to the 1024-token position distribution seen at fit. The
decoupled formulation extends to longer contexts by construction, which
matters for serving prompts up to 32k tokens (\S\ref{sec:latency}).

Concretely, we strip source RoPE, apply $\mathbf{W}_K$ in the position-free
space, and re-encode with target RoPE:
$\hat{\mathbf{K}}_t = (\mathbf{K}_s\,\mathbf{R}_{\Theta_s}^{-1}(t)\,\mathbf{W}_K + \mathbf{b}_K)\,\mathbf{R}_{\Theta_t}(t)$.
During calibration the regression targets $\mathbf{Y}$ are constructed by
stripping the target RoPE from the target's ground-truth keys, so $\mathbf{W}_K$
is fit entirely in the position-free space.
Since $\mathbf{R}_{\Theta}$ is orthogonal, the inversion is exact at negligible cost.
Values carry no positional encoding and are mapped directly.

\vspace{-0.5em}
\section{Experiments}
\label{sec:experiments}

\vspace{-0.5em}
We evaluate four questions: (1)~How well does ridge match the target's own KV
across three families (\S\ref{sec:main})? (2)~Which components matter
(\S\ref{sec:ablation})? (3)~Can a nonlinear mapper recover the pairs where
ridge falls short (\S\ref{sec:nonlin}), and what determines per-pair outcomes
(\S\ref{sec:mechanism})? (4)~Does transfer hold up in multi-turn
(\S\ref{sec:multiturn}), and what is the latency advantage
(\S\ref{sec:latency})?

\vspace{-5pt}
\subsection{Setup}
\label{sec:setup}

\vspace{-5pt}
\paragraph{Model families.} We evaluate three matched-KV families
(Qwen3, Llama 3.1, Ministral 3) where source and target share KV head
count and per-head dimension across scales (full table in
Appendix~\ref{app:setup}). All families use dense full-attention, so
every target layer receives mapped KV. Models are post-trained (Qwen3,
Ministral 3) or base (Llama 3.1), evaluated in completion mode.

\vspace{-5pt}
\paragraph{Benchmarks.} We evaluate on five accuracy benchmarks plus
WikiText-2 perplexity: ARC-Challenge \citep{clark2018arc}, HellaSwag
\citep{zellers2019hellaswag}, WinoGrande \citep{sakaguchi2020winogrande},
MMLU 5-shot \citep{hendrycks2021mmlu}, GSM8K 8-shot with chain-of-thought
prompting \citep{cobbe2021gsm8k,wei2022cot}, and prefix-conditioned
WikiText-2 perplexity.
CoQA \citep{reddy2019coqa} measures multi-turn handoff (\S\ref{sec:multiturn}).
All five accuracy benchmarks are evaluated on all six pairs at each pair's
selected $k$. Perplexity coverage and its protocol are in
Appendix~\ref{app:setup}. Retention is
$(\text{transfer accuracy}) / (\text{target standalone accuracy}) \times 100\%$.
Benchmarks differ in their chance floor, so we also report floor-normalized
retention, $(\text{acc}-\text{chance})/(\text{target}-\text{chance})$, which
places chance at 0\% and the target's own accuracy at 100\% on every
benchmark.

\vspace{-5pt}
\paragraph{Mapper configuration.} All pairs use ridge ($\lambda=0.01$) and
content-space mapping. The number of source layers $k$ is swept over
$\{1,2,4,6,8,10,12,16,20,24,\text{all}\}$. We \emph{select} a single $k$
per pair as the value that maximizes the arithmetic mean of the
log-likelihood benchmark accuracies (ARC-C, HellaSwag, WinoGrande, MMLU),
breaking near-ties toward larger $k$ for evaluation coverage.
GSM8K, CoQA multi-turn, and prefill latency are held out from this
selection. Including a benchmark in the criterion raises its own reported
accuracy by at most $2.49$\,pp (Appendix~\ref{app:k_selection}).
Per-pair selected $k$, total mapper parameters
(1.01--3.36\,B), storage (4--12\,GB), and the family table are in
Appendix~\ref{app:setup}.

\vspace{-5pt}
\subsection{Main results}
\label{sec:main}
\vspace{-5pt}
Main results focus on small-to-large (S$\to$L) transfer across six
matched-KV pairs from three families. Large-to-small (L$\to$S)
evaluation is limited to HellaSwag and appears in the mechanism analysis
(\S\ref{sec:mechanism}) and multi-turn evaluation (\S\ref{sec:multiturn}).
All pairs are evaluated at their selected~$k$ with the full pipeline.
Outcomes vary widely (42--98\% Avg retention). Table~\ref{tab:main_results}
reports the headline per-pair numbers. Full per-family tables are in
Appendix~\ref{app:extended_results}.

\begin{table}[t]
  \caption{Per-pair transfer retention at the selected $k$ (ridge,
  content-space) across six matched-KV pairs from three families.
  Per-benchmark columns are retention, with each benchmark's chance floor in
  the first row. Avg and Avg$_{\text{fn}}$ are the means of retention and of
  floor-normalized retention across those benchmarks, computed from unrounded
  values. Per-benchmark floor-normalized values are in
  Appendix~\ref{app:extended_results}.}
  \label{tab:main_results}
  \centering
  \small
  \setlength{\tabcolsep}{3.5pt}
  \vspace{-0.3em}
  \begin{tabular}{llrrrrrrr}
    \toprule
    \multicolumn{1}{c}{Family} & \multicolumn{1}{c}{Pair ($k$)} & \multicolumn{1}{c}{Avg} & \multicolumn{1}{c}{Avg$_{\text{fn}}$} & \multicolumn{1}{c}{ARC-C} & \multicolumn{1}{c}{HellaSwag} & \multicolumn{1}{c}{WinoGrande} & \multicolumn{1}{c}{MMLU} & \multicolumn{1}{c}{GSM8K} \\
    \midrule
    \multicolumn{2}{l}{\emph{Chance floor}} & --- & --- & 25\% & 25\% & 50\% & 25\% & $\approx$0\% \\
    \midrule
    Qwen3       & 14B$\to$32B (8)   & 97.6\% & 96.3\% & 101.0\% & 97.6\% & 98.5\% & 95.0\% & 95.6\% \\
    Qwen3       & 8B$\to$32B (12)   & 87.5\% & 80.7\% &  94.0\% & 95.2\% & 91.0\% & 88.5\% & 68.8\% \\
    Llama 3.1   & 8B$\to$70B (20)   & 72.8\% & 62.9\% &  90.9\% & 94.4\% & 87.1\% & 73.3\% & 18.2\% \\
    Ministral 3 & 3B$\to$8B (all)   & 76.2\% & 65.9\% &  90.6\% & 93.3\% & 91.3\% & 69.4\% & 36.6\% \\
    Ministral 3 & 3B$\to$14B (20)   & 44.2\% & 14.7\% &  43.6\% & 68.0\% & 74.0\% & 32.0\% &  3.2\% \\
    Ministral 3 & 8B$\to$14B (12)   & 41.6\% & 11.1\% &  40.7\% & 58.7\% & 74.2\% & 32.7\% &  1.6\% \\
    \bottomrule
  \end{tabular}
\end{table}

Two observations structure the rest of the paper.
\textbf{(1) Tier 1.} Four matched-KV pairs retain $73$--$98\%$ of target
accuracy averaged across the five benchmarks: Qwen3 14B$\to$32B, Qwen3 8B$\to$32B, Llama 3.1 8B$\to$70B
(the most extreme parameter ratio in our evaluation), and Ministral
3B$\to$8B.
\textbf{(2) Tier 2.} Two matched-KV pairs degrade sharply:
Ministral 8B$\to$14B and Ministral 3B$\to$14B fall to 42--44\% Avg, and to
11--15\% once floor-normalized.
Together, the two tiers show that matched KV correlates with success
but does not guarantee it. \S\ref{sec:nonlin} and \S\ref{sec:mechanism}
investigate what additional factors determine per-pair outcomes.
Figure~\ref{fig:retention} (Appendix~\ref{app:extended_results}) visualizes
retention across benchmarks.

\vspace{-0.3em}
\subsection{Ablation}
\label{sec:ablation}

\begin{table}[t]
  \caption{Mapper component ablation on Qwen3 14B$\to$32B. Cells are
  accuracy per benchmark, with WikiText-2 perplexity in the rightmost
  column. ``Full'' uses $k{=}8$, ridge, and content-space (RoPE-stripped)
  mapping. ``$-$ inference RoPE'' applies the content-space fit without
  re-rotating at inference, creating a fit-vs-eval mismatch. ``$-$ all
  RoPE'' is the fully coupled variant (RoPE retained at both fit and
  inference). The last two rows additionally drop cross-layer source
  selection ($k{=}1$) and ridge regularization.}
  \label{tab:ablation}
  \centering
  \small
  \setlength{\tabcolsep}{4pt}
  \begin{tabular}{lrrrrrr}
    \toprule
    \multicolumn{1}{c}{Configuration} & \multicolumn{1}{c}{ARC-C} & \multicolumn{1}{c}{HellaSwag} & \multicolumn{1}{c}{WinoGrande} & \multicolumn{1}{c}{MMLU} & \multicolumn{1}{c}{GSM8K} & \multicolumn{1}{c}{PPL} \\
    \midrule
    Full ($k{=}8$, ridge, content-space) & 61.60 & 80.70 & 68.98 & 78.09 & 90.98 & 7.33 \\
    $-$ inference RoPE                   & 44.97 & 75.39 & 56.59 & 25.79 &  4.17 & 7.70 \\
    $-$ all RoPE (fit + inference)       & 61.09 & 80.73 & 68.59 & 77.70 & 90.98 & 7.35 \\
    $-$ RoPE $-$ cross-layer ($k{=}1$)   & 27.65 & 44.81 & 51.78 & 26.07 &  0.38 & 22.73 \\
    $-$ RoPE $-$ cross-layer $-$ ridge   & 36.43 & 62.26 & 51.22 & 51.26 &  1.44 &  9.86 \\
    \bottomrule
  \end{tabular}
\end{table}

\vspace{-0.3em}
We characterize the mapper on Qwen3 14B$\to$32B, the highest-retention pair, where
any degradation is most visible. Table~\ref{tab:ablation} reports
component ablations, and calibration sensitivity to
$\lambda$, $N$, and calibration domain is in Appendix~\ref{app:ablation}.

\textbf{(1) Cross-layer source selection is the largest contributor.}
Reducing $k$ from $8$ to $1$ drops K $R^2$ from $0.79$ to $0.56$
(Appendix~\ref{app:ablation}).
\textbf{(2) Content-space mapping is benchmark-specific.} Disabling
inference-time RoPE handling collapses MMLU and GSM8K to near random while
HellaSwag falls only $\sim$5\,pp. \textbf{(3) Calibration is robust.} A
four-order-of-magnitude sweep of $\lambda$ and a 50--1000 sequence sweep of
$N$ both show wide flat regions, with collapse only at $\lambda{=}1$. Sample
count flattens after $N{=}200$, with $N{=}50$ still within $\sim$1.6\,pp of
production. Domain is the one axis with real cost: CodeAlpaca drops
5.24\,pp on HellaSwag while Wikipedia stays within noise.

\vspace{-4pt}
\subsection{Substituting MLP for ridge}
\label{sec:nonlin}

The natural follow-up to \S\ref{sec:main} is whether a nonlinear mapper
recovers the pairs where ridge falls short. We train a per-(target layer,
head, K$|$V) MLP on the same MSE loss as ridge and use it as a drop-in
replacement at inference. The MLP has two 1{,}024-unit ReLU-activated hidden layers
trained with Adam (full settings in Appendix~\ref{app:run_info}). We
evaluate four Qwen3 and Ministral 3 pairs covering the success-to-failure
range using the same downstream evaluation pipeline as ridge, so the only
factor that changes is the mapper functional form.

Table~\ref{tab:mlp_vs_ridge} shows two observations. On the pairs where
ridge already succeeds, MLP slightly underperforms ridge. On the pairs
where ridge fails, MLP recovers HellaSwag retention by $+24.3$ to
$+36.8$\,pp, which puts all four pairs above $90\%$ under the MLP.
Linear ridge is sufficient where the cross-model KV
relationship is already linear, and the MLP helps only where ridge falls
short rather than dominating ridge across the board.

\begin{table}[t]
  \caption{Ridge vs.\ MLP HellaSwag retention on four Qwen3 and Ministral
  3 pairs.}
  \label{tab:mlp_vs_ridge}
  \centering
  \small
  \setlength{\tabcolsep}{8pt}
  \begin{tabular}{lrrr}
    \toprule
    \multicolumn{1}{c}{Pair} & \multicolumn{1}{c}{Ridge} & \multicolumn{1}{c}{MLP} & \multicolumn{1}{c}{$\Delta$} \\
    \midrule
    Qwen3 14B$\to$32B    & 97.6\% & 97.3\% & $-0.3$\,pp \\
    Ministral 3B$\to$8B  & 93.3\% & 91.8\% & $-1.5$\,pp \\
    \textbf{Ministral 3B$\to$14B} & \textbf{68.0\%} & \textbf{92.3\%} & $\bm{+24.3}$\,\textbf{pp} \\
    \textbf{Ministral 8B$\to$14B} & \textbf{58.7\%} & \textbf{95.5\%} & $\bm{+36.8}$\,\textbf{pp} \\
    \bottomrule
  \end{tabular}
\end{table}

\textbf{What changes between ridge and MLP on the failure pairs?} On the
HellaSwag tokens used for evaluation, ridge's $R^2_K$ is deeply negative
(Table~\ref{tab:ridge_vs_mlp_conc}), meaning the calibration-fit linear
mapper does not extrapolate. MLP closes most of that gap.
\S\ref{sec:mechanism} shows that calibration $R^2$ does not predict
retention across pairs, and that this gain coincides with lower error
concentration and higher attention-output cosine.

\vspace{-4pt}
\subsection{What determines transfer quality?}
\label{sec:mechanism}

Calibration $R^2$ is the natural a priori metric for transfer quality.
If it predicted retention, deployment screening could rely on the fit
alone. Across our six matched-KV pairs evaluated in both directions,
however, $R^2$ alone does not predict per-pair outcomes. Llama 3.1
8B$\to$70B fits ridge with $R^2_K{=}0.84$ on calibration data and retains
94\% HellaSwag small-to-large, but only 37\% large-to-small. Ministral
3B$\to$8B fits at the same $R^2_K{=}0.84$ and retains 93\% in both
directions. The same calibration $R^2$ produces very different downstream
outcomes. \S\ref{sec:nonlin} additionally showed that switching ridge for
an MLP recovers up to $+37$\,pp downstream on the pairs where ridge
falls short. Two questions follow. What scalar
predicts retention better than $R^2$ across pairs, and what does the MLP
do differently from ridge on the failure pairs?

\paragraph{Attention-output cosine.}
$R^2$ measures how closely the mapper reconstructs each K and V channel
on average, weighting all dimensions equally. Attention does not. It
scores K against the target's queries $\mathbf{Q}$ and weights V by the
resulting attention pattern. The quantity that decides whether downstream
behavior is preserved is the \emph{attention output} the target would have
computed. We measure it directly with cosine similarity between the
attention output from mapped KV and from ground-truth KV, averaged over
layers and heads. Across 12 matched-KV pair evaluations from three
families (six S$\to$L and six L$\to$S directions), mean cosine
correlates with HellaSwag retention at Pearson $r{=}{+}0.57$, while
calibration-domain $R^2_K$ shows essentially no correlation
($r{=}{-}0.20$). $R^2$ remains useful within a single pair, for example
for source-layer selection, but is not the right scalar across pairs.

Cosine summarizes attention fidelity at the pair level. To explain
\emph{why} different mappers with similar $R^2$ produce different cosine,
we look at where the residual error lands per head.

\paragraph{Where the mapper's error lands.}
We measure \emph{error concentration} per head. For K, we project the
mapper's per-token K error onto the right singular vectors of the
target's per-head query matrix $\mathbf{Q}_h$ and weight each component
by the matching squared singular value. Dividing by the mean error across
all components gives the K-concentration. For V, we weight per-position
V error by the squared GT attention weight at that position and divide by
the mean per-position error. Both quantities are averaged across heads.
Concentration above 1 means the error concentrates where attention reads,
while below 1 means it lands where attention ignores.

\paragraph{MLP intervention.}
This explains the \S\ref{sec:nonlin} puzzle and clarifies the role of
the nonlinear mapper. We compare ridge and MLP on the same four pairs,
same calibration data, only the mapper changing.
Table~\ref{tab:ridge_vs_mlp_conc} shows the per-pair shifts. On the pairs
where ridge falls short the MLP substantially lowers K-concentration
($\Delta{\approx}{-}2.5$ on average) and raises attention-output cosine
($\Delta{\approx}{+}0.45$), with HellaSwag retention rising by $+24.3$ to
$+36.8$\,pp. Where ridge already succeeds the shifts are far smaller and retention does
not follow: Ministral 3B$\to$8B improves on both quantities and still loses
ground on HellaSwag. Redistributing error
is therefore not sufficient on its own. It changes downstream accuracy
only where the misplaced error was large enough to bind, which is why
linear ridge is enough wherever the cross-model KV relationship is already
linear. The same shift lifts eval-domain $R^2_K$ from deeply negative to
near zero on the failure pairs, though it stays below zero: ridge's
calibration-domain fit does not extrapolate to HellaSwag tokens, while the
MLP closes most of that gap.

\begin{table}[t]
  \caption{Per-pair shifts when ridge is replaced by MLP
  ($\Delta = \text{MLP} - \text{ridge}$, all measured on HellaSwag
  tokens). HS values are HellaSwag retention from
  Table~\ref{tab:mlp_vs_ridge}. Ridge's $R^2_K$ is deeply negative on
  the failure pairs, indicating the calibration-fit linear mapper does
  not extrapolate. On the bolded failure pairs MLP lifts $R^2_K$ toward
  zero, lowers K-concentration, raises attention-output cosine, and
  recovers HellaSwag retention.}
  \label{tab:ridge_vs_mlp_conc}
  \centering
  \small
  \resizebox{\linewidth}{!}{%
  \begin{tabular}{lrrrrrrrrr}
    \toprule
    Pair & Ridge $R^2_K$ & $\Delta R^2_K$ & Ridge $R^2_V$ & $\Delta R^2_V$ & Ridge HS & $\Delta$ HS & $\Delta$ K-conc & $\Delta$ V-conc & $\Delta$ cosine \\
    \midrule
    Qwen3 14B$\to$32B    & 0.75 & $-0.05$ & 0.56 & $-0.06$ & 97.6\% & $-0.3$\,pp & $-0.03$ & $+0.00$ & $-0.03$ \\
    Ministral 3B$\to$8B  & $-0.10$ & $+0.77$ & 0.59 & $-0.09$ & 93.3\% & $-1.5$\,pp & $-0.45$ & $-0.08$ & $+0.07$ \\
    \textbf{Ministral 3B$\to$14B} & $\bm{-7.81}$ & $\bm{+7.62}$ & \textbf{0.17} & $\bm{+0.27}$ & \textbf{68.0\%} & $\bm{+24.3}$\,\textbf{pp} & $\bm{-2.31}$ & $\bm{-0.23}$ & $\bm{+0.41}$ \\
    \textbf{Ministral 8B$\to$14B} & $\bm{-3.22}$ & $\bm{+3.08}$ & \textbf{0.19} & $\bm{+0.37}$ & \textbf{58.7\%} & $\bm{+36.8}$\,\textbf{pp} & $\bm{-2.71}$ & $\bm{-0.23}$ & $\bm{+0.48}$ \\
    \bottomrule
  \end{tabular}}
\end{table}

\vspace{-4pt}
\subsection{Multi-turn handoff}
\label{sec:multiturn}

\begin{figure}[t]
  \centering
  \IfFileExists{figures/retention_vs_turns.png}
    {\includegraphics[width=0.85\linewidth]{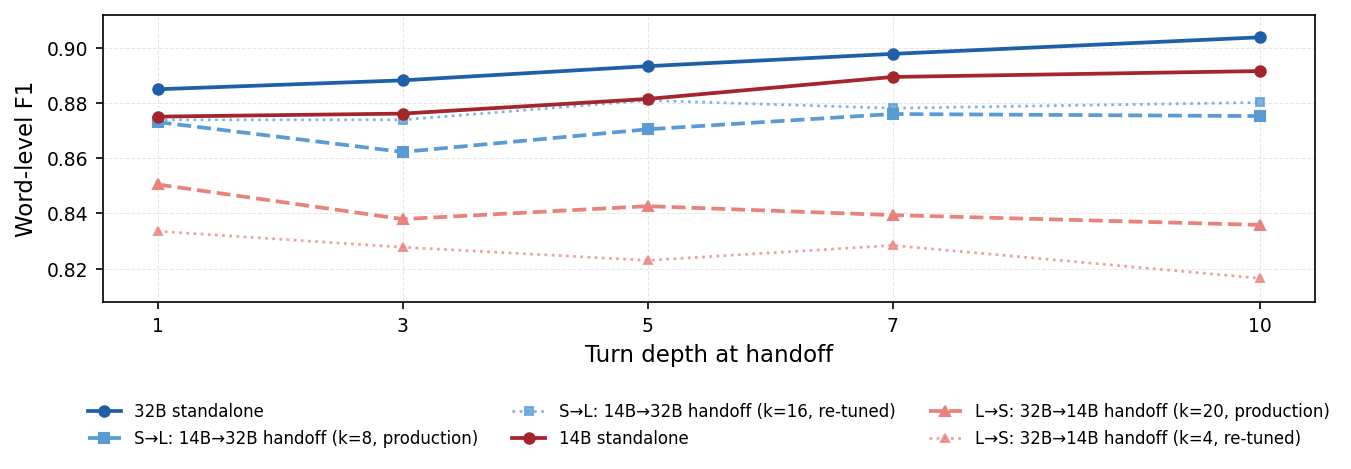}}
    {\fbox{\rule{\linewidth}{0pt}\rule{0pt}{4cm}}}
  \caption{CoQA F1 across ten turns for Qwen3 14B$\leftrightarrow$32B
  multi-turn handoff. Solid lines are the target's standalone accuracy,
  dashed lines the mapper at the single-turn $k$ of \S\ref{sec:setup},
  $k{=}8$ (S$\to$L) and $k{=}20$ (L$\to$S), and dotted lines the $k$ a
  multi-turn sweep would select.}
  \label{fig:retention_vs_turns}
\end{figure}

The mid-conversation switching scenario alternates between source and
target across turns, so we measure drift on Qwen3 14B$\leftrightarrow$32B
with CoQA \citep{reddy2019coqa} on 100 conversations of $\sim$15 turns
each across five domains. We score each turn's answer
with F1 against the ground-truth and define \emph{drift} at turn $t$ as
the F1 gap between the target's standalone accuracy and the mapper's
accuracy at the same turn.
Drift stays small in both directions
(Figure~\ref{fig:retention_vs_turns}). The small-to-large gap widens by
$1.7$\,pp from turn 1 to turn 10, with the mapper holding steady while the
32B ceiling rises. Large-to-small drift grows linearly at $0.33$\,pp/turn. Both slopes are too
small for cascading
failure within ten turns on this single-pair evaluation, though linear
large-to-small drift would still accumulate over very long sessions. Tuning
$k$ on the multi-turn task changes drift by at most $2.0$\,pp (dotted lines).

\vspace{-4pt}
\subsection{Prefill latency}
\label{sec:latency}

\begin{table}[t]
  \caption{Re-prefill vs.\ KV cache transfer latency on Qwen3
  14B$\leftrightarrow$32B at each direction's selected $k$, $k{=}8$ for
  S$\to$L and $k{=}20$ for L$\to$S.}
  \label{tab:latency}
  \centering
  \small
  \setlength{\tabcolsep}{5pt}
  \begin{tabular}{lrrrrrr}
    \toprule
     & \multicolumn{3}{c}{S$\to$L (14B$\to$32B)} & \multicolumn{3}{c}{L$\to$S (32B$\to$14B)} \\
    \cmidrule(lr){2-4} \cmidrule(lr){5-7}
    Seq len & Mapper (ms) & Re-prefill (ms) & Speedup & Mapper (ms) & Re-prefill (ms) & Speedup \\
    \midrule
    64    & 14.0   & 61.7    & 4$\times$ & 11.6   & 39.2    & 3$\times$ \\
    8K    & 67.8   & 1154.8  & 17$\times$ & 101.9  & 501.0   & 5$\times$ \\
    32K   & 277.6  & 6975.3  & 25$\times$ & 427.1  & 2952.7  & 7$\times$ \\
    \bottomrule
  \end{tabular}
\end{table}

The cost we want to skip is the receiver's prefill, so we compare end-to-end
re-prefill latency against KV cache transfer latency on Qwen3 14B$\leftrightarrow$32B.
The mapper replaces the target's transformer body with a per-layer batched
matrix multiply, letting the receiver decode directly from the mapped cache.
Table~\ref{tab:latency} reports per-sequence-length latency. The mapper runs
4--25$\times$ faster small-to-large and 3--7$\times$ faster large-to-small
across sequence lengths from 64 to 32{,}768 tokens. The large-to-small
speedup widens with sequence length because the mapper's wall time grows
much slower than the receiver's re-prefill. The full sequence-length
sweep is in Appendix~\ref{app:latency}.

\section{Broader impact, future work, and limitations}
\label{sec:limitations}

\paragraph{Broader impact.}
Cross-model KV cache transfer skips re-prefill on the receiver,
reducing serving cost, energy, and tail latency for agentic workloads
with frequent model swaps in cost-quality cascading,
mid-conversation switching, and routing.
\vspace{-5pt}

\paragraph{Future work.}
(1)~\textbf{Trained mapper alternatives.} Linear regression fits from
a small calibration set without backpropagation, which made it the
practical starting point. The MLP results in \S\ref{sec:nonlin} show
that nonlinear variants can recover the matched-KV pairs where ridge
falls short, so systematic comparison across MLP, MoE, and
attention-aligned objectives that directly optimize cosine is a
natural follow-up.
(2)~\textbf{Predictive transferability.} Attention-output cosine acts
as a post-hoc diagnostic since it requires fitting the mapper. A
signal that estimates transferability before fitting would
accelerate deployment screening, with downstream retention on
representative workloads such as HellaSwag as one candidate proxy.
(3)~\textbf{Cross-family transfer.} Whether the closed-form approach
extends across families such as Qwen3$\to$Llama 3.1 is open. Different
families may share enough representation structure to admit a similar
mapping, or may require different machinery.
(4)~\textbf{Hidden factors.} Beyond architecture, training-data
overlap, fine-tuning recipe, and other non-architectural factors may
also influence transferability. We leave systematic study of these to
future work.
(5)~\textbf{Hybrid architectures.} Extending the mapper to
attention-restricted hybrids (sliding-window, local) and to
attention-recurrent hybrids such as Nemotron 3
\citep{blakeman2025nvidia} that carry SSM state alongside KV is
a natural next step.
\vspace{-5pt}

\paragraph{Limitations.}
(1)~\textbf{Single-domain calibration.} Calibration uses only FineWeb-Edu.
Appendix~\ref{app:ablation} measures the cost of substituting Wikipedia or
CodeAlpaca on one pair. Neither substitution separates subject matter from
register, so the sweep does not bound calibration confined to a single field
such as medicine or law.
(2)~\textbf{$k$ selection.} Per-pair $k$ is selected on the same
log-likelihood benchmarks we report on. Appendix~\ref{app:k_selection}
measures the effect, at most $2.49$\,pp, and adds a held-out evaluation,
though neither is a substitute for selecting $k$ out of sample.
(3)~\textbf{Matched-KV is empirical.} The closed-form mapper imposes
no structural requirement on source or target dimensions, but our
six pairs are all matched-KV by construction. We do not test
mismatched-KV pairs. Whether ridge transfer can succeed under
mismatched-KV conditions is left to future work.
(4)~\textbf{Scope.} Within-family transfer over dense full-attention
models. Hybrid attention and attention-recurrent architectures are
out of scope.
\vspace{-5pt}

\section{Conclusion}
\label{sec:conclusion}
Cross-model KV cache transfer within an LLM family admits a closed-form,
training-free fit. The cross-model KV relationship is largely linear,
and a closed-form per-head ridge mapping retains 73--98\% of standalone
accuracy on the four highest-retention matched-KV pairs in our
evaluation, runs 2.7--25$\times$ faster than re-prefill, and remains stable
across multi-turn handoff. Per-pair retention is determined by where the
residual error lands relative to the target's attention-sensitive
subspaces, not by its magnitude.
Attention-output cosine captures this
and predicts cross-pair retention better than $R^2$ ($r{=}{+}0.57$
vs.\ $r{=}{-}0.20$ over 12 matched-KV pair evaluations from three
families), and a nonlinear MLP adds up to $+37$\,pp HellaSwag retention
on harder pairs by redistributing error toward attention-irrelevant
directions. These results shift evaluation criteria for cross-model
mappers from raw reconstruction metrics to subspace-aware diagnostics,
and motivate attention-aligned objectives for future cross-model mappers.


\bibliographystyle{plainnat}
\bibliography{references}

\newpage

\appendix
\raggedbottom

\section{Cross-model KV cache transfer methods}
\label{app:prior_comparison}

This appendix tabulates the cross-model KV cache transfer methods discussed
in the prose comparison of \S\ref{sec:prior}. Each row records whether the
method is gradient-free, whether it works cross-scale (different parameter
counts), whether it transfers KV values rather than attention patterns, and
whether the mapping is closed-form. Our method is the only one that
satisfies all four criteria.

\begin{table}[H]
  \caption{Comparison of cross-model KV cache transfer methods.}
  \label{tab:prior}
  \centering
  \small
  \begin{tabular}{lcccc}
    \toprule
    Method & Gradient-free & Cross-scale & Transfers KV values & Closed-form \\
    \midrule
    C2C \citep{fu2026c2c}             & --            & \checkmark & \checkmark & --         \\
    LatentAlign \citep{dery2026latentalign} & --      & \checkmark & \checkmark & --         \\
    IAM \citep{zhao2025iam}           & \checkmark    & \checkmark & --         & --         \\
    DroidSpeak \citep{liu2026droidspeak}\textsuperscript{$\dagger$} & \checkmark & --       & \checkmark & --         \\
    \textbf{Ours}                     & \textbf{\checkmark} & \textbf{\checkmark} & \textbf{\checkmark} & \textbf{\checkmark} \\
    \bottomrule
  \end{tabular}\\[2pt]
  \footnotesize $^\dagger$\,DroidSpeak transfers KV only between architecturally
  identical models (same hidden size, layer count, head configuration). The
  cross-scale and closed-form columns are therefore marked ``---'' as
  inapplicable rather than absent.
\end{table}

\section{Linear structure analysis}
\label{app:linear_structure}

This appendix collects the empirical evidence for the linear-structure claim in
\S\ref{sec:linear_structure}: the greedy forward-selection curves that
quantify how many source layers the mapper actually needs. The pairwise $R^2$
heatmaps that motivate the four qualitative patterns are inlined as
Figure~\ref{fig:r2-heatmaps} in \S\ref{sec:linear_structure}, and the
production-ridge $R^2$ at $k{=}1, 8, \text{all}$ is in
Appendix~\ref{app:ablation}.

\paragraph{Source layer selection by greedy forward selection.}
The single-source $R^2$ probe in \S\ref{sec:linear_structure} measures how
predictable each target layer is from a single source layer. Here we ask how
many source layers the mapper actually needs by greedy forward selection.
Starting from the best single source layer, we iteratively add the source
layer that maximally increases joint $R^2$ at each step, producing curves
from $k{=}1$ to $k{=}\text{all}$.

\begin{figure}[H]
  \centering
  \IfFileExists{figures/concentration_comparison.png}
    {\includegraphics[width=\linewidth]{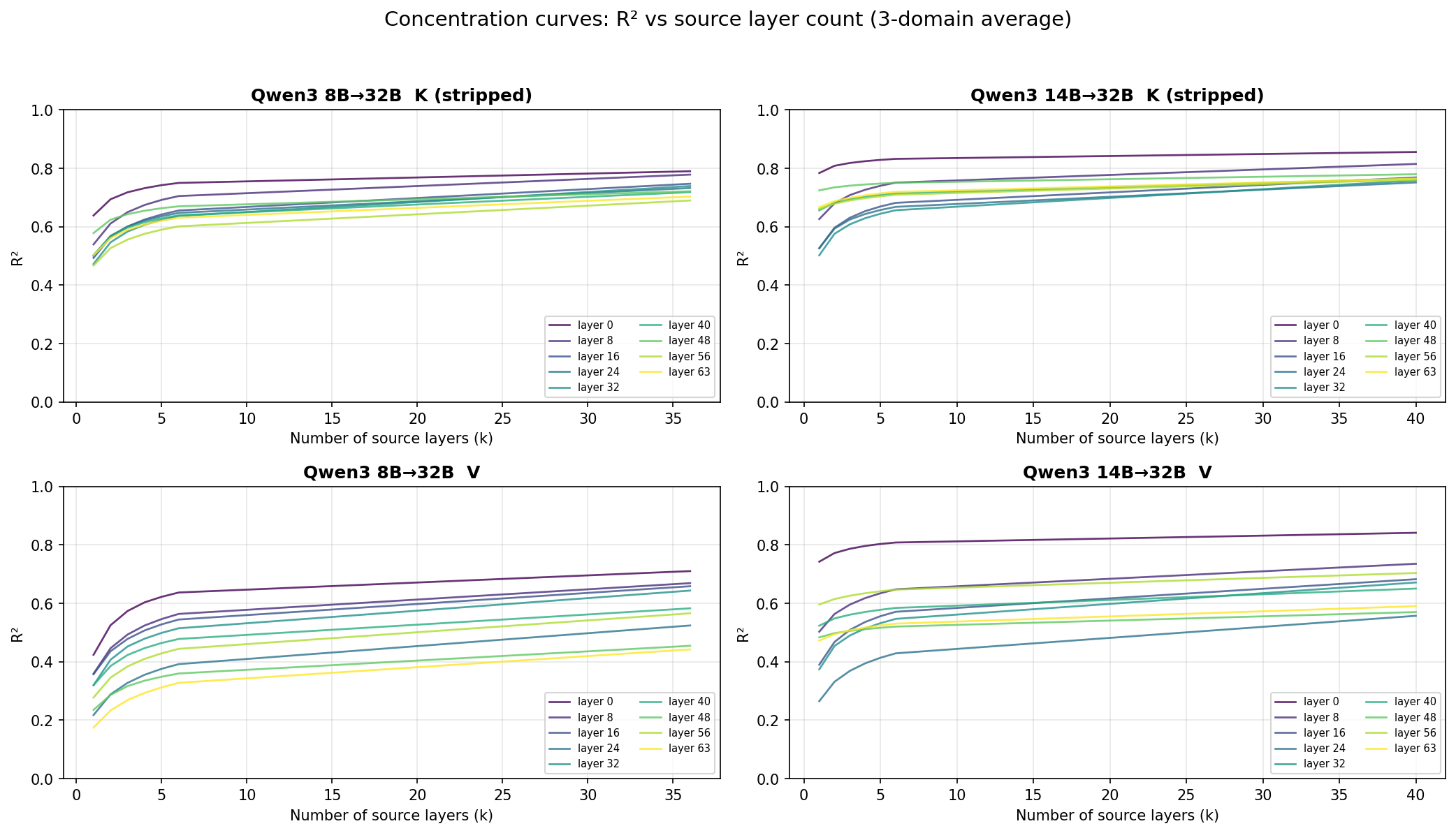}}
    {\fbox{\rule{\linewidth}{0pt}\rule{0pt}{4cm}}}
  \caption{$R^2$ vs.\ number of source layers selected via greedy forward
  selection, for $K_{\text{stripped}}$ and $V$ (rows) and
  Qwen3 8B$\to$32B, 14B$\to$32B (columns). Each curve is one of nine
  representative target layers. Curves climb steeply from $k{=}1$ to $k{=}4$
  and are close to their $k{=}\text{all}$ value by $k{=}6$. For Qwen3
  14B$\to$32B, $R^2$ at $k{=}6$ reaches
  $92.3\%$ of the $k{=}\text{all}$ value for $K_{\text{stripped}}$ and $87.7\%$
  for $V$.}
  \label{fig:concentration}
\end{figure}

The curves climb steeply from $k{=}1$ to $k{=}4$ and are close to their
$k{=}\text{all}$ value by $k{=}6$.
Quantitatively, even on the sharpest pair (Qwen3 14B$\to$32B), picking the
best single source layer per target and averaging across target layers reaches
only $R^2{=}0.56$ for $K_{\text{stripped}}$ and $0.32$ for $V$, while $k{=}8$
aggregation reaches $0.79$ and $0.65$ respectively. The same $0.56$ appears as
the production-ridge $k{=}1$ row of Table~\ref{tab:k_sweep_r2}; this is a
weaker quantity than the per-cell heatmap peak of $0.81$ in
\S\ref{sec:linear_structure}, which singles out the most predictable
target layer. Complementary information is distributed across layers for both
K and V, motivating top-$k$ cross-layer selection in our mapper design
(\S\ref{sec:topk}). Figure~\ref{fig:concentration} uses greedy forward
selection (joint $R^2$ maximization per target layer and head). Our production
mapper (\S\ref{sec:topk}) uses fixed top-$k$ by single-source head-averaged
$R^2$ for tractability. Both support the same qualitative conclusion that
complementary information is distributed across multiple source layers.

\section{Extended ablation}
\label{app:ablation}

This appendix collects the ablation tables and figures supporting
\S\ref{sec:ablation}: the source layer count sweep, the full
sequential-removal table, and calibration sensitivity to ridge $\lambda$,
sample count $N$, and domain.

\paragraph{Source layer count.}
The fit $R^2$ increases with $k$ but with diminishing returns:

\begin{table}[H]
  \caption{K and V in-sample $R^2$ vs.\ source layer count $k$ on Qwen3
  14B$\to$32B.}
  \label{tab:k_sweep_r2}
  \centering
  \small
  \begin{tabular}{lrr}
    \toprule
    $k$ & K $R^2$ (14B$\to$32B) & V $R^2$ (14B$\to$32B) \\
    \midrule
    1   & 0.5572 & 0.3249 \\
    8   & 0.7914 & 0.6541 \\
    all & 0.8451 & 0.7645 \\
    \bottomrule
  \end{tabular}
\end{table}

Going from $k{=}1$ to $k{=}8$ provides the largest K $R^2$ gain. $k{=}8$ to
$k{=}\text{all}$ yields diminishing returns.

Downstream, the same saturation pattern holds across pairs and benchmarks
(Figure~\ref{fig:k-sweep}). $k{=}1$ is uniformly insufficient. Every pair
drops sharply on at least one benchmark. Closer pairs saturate earlier.
Qwen3 14B$\to$32B reaches within $0.3$\,pp of peak HellaSwag by $k{=}8$,
while the more distant Llama 3.1 8B$\to$70B keeps improving until
$k{=}24$.

\IfFileExists{figures/accuracy_vs_k.png}
{\begin{figure}[H]
  \centering
  \scalebox{1}[0.85]{\includegraphics[width=\linewidth]{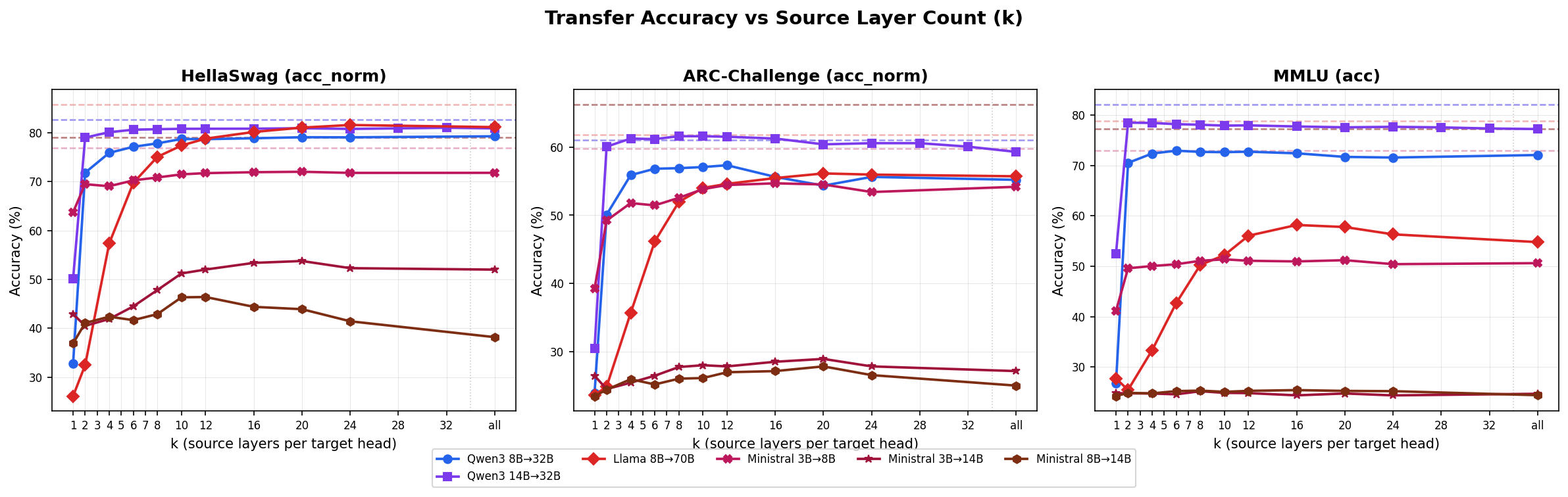}}
  \caption{Transfer accuracy vs.\ source layer count $k$ across benchmarks
  (panels) for within-family pairs (curves). Dashed lines mark target
  standalone. Sweep:
  $k\in\{1,2,4,6,8,10,12,16,20,24,\text{all}\}$.}
  \label{fig:k-sweep}
\end{figure}}{}

\paragraph{Sequential component removal.}
Removing each mapper component in turn isolates its contribution
(Figure~\ref{fig:seq-removal} and Table~\ref{tab:ablation}). \S\ref{sec:ablation}
already highlights cross-layer source selection ($k{=}8\to k{=}1$) as the
largest contributor. Fitting and inferring on RoPE-coupled keys (the
``$-$ all RoPE'' row) lands within noise of the full decoupled pipeline on
every benchmark at the 1{,}024-token fit context. The decoupled formulation
is preferred because it generalizes across RoPE configurations and to longer
contexts by construction (\S\ref{sec:rope}), not because the coupled variant
fails on the calibration distribution.

\IfFileExists{figures/sequential_removal.png}
{\begin{figure}[H]
  \centering
  \scalebox{1}[0.78]{\includegraphics[width=\linewidth]{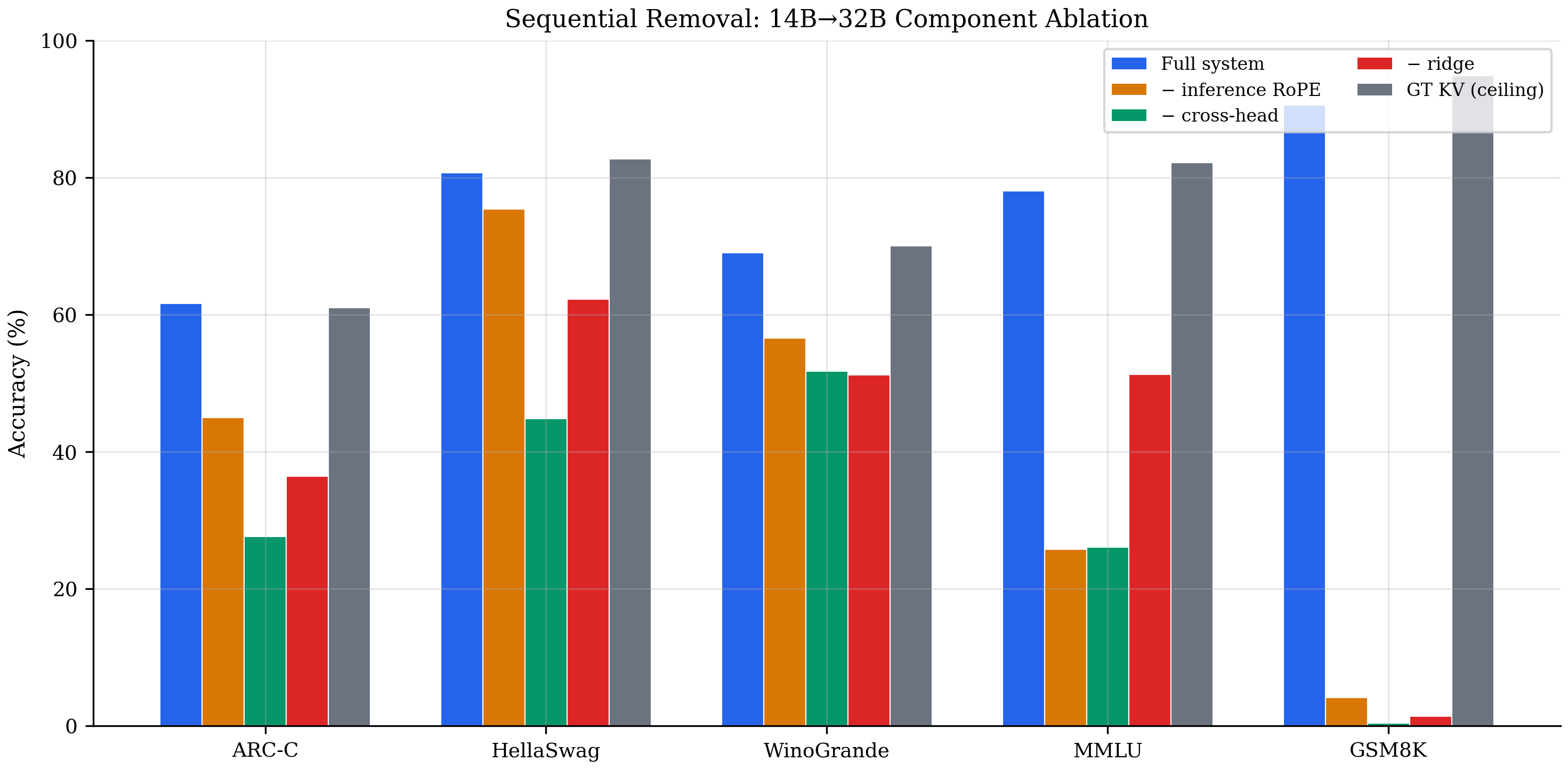}}
  \caption{Sequential removal of mapper components on Qwen3 14B$\to$32B, with
  the target's own ground-truth cache as the ceiling. Removing cross-layer
  source selection (green; $k{=}1$) costs the most on ARC-C and HellaSwag,
  while removing inference-time RoPE handling costs the most on WinoGrande,
  MMLU and GSM8K. Numerical values in Table~\ref{tab:ablation}.}
  \label{fig:seq-removal}
\end{figure}}{}

The numerical sequential-removal table is in the body
(Table~\ref{tab:ablation}, \S\ref{sec:ablation}).

\paragraph{Calibration protocol sensitivity.}
Both ridge $\lambda$ (four orders of magnitude around the production $0.01$) and
sample count $N$ (50--1000 sequences) have wide flat regions around the production
value, because the per-head system is over-determined. Collapse appears only at
extremes. $\lambda{=}1$ ($-15.79$\,pp on HellaSwag) drives the ridge penalty past
the least-squares objective. $N$ flattens after 200 sequences, with $N{=}50$ still
within $\sim$1.6\,pp of production. Domain is the one axis with a real cost
(Table~\ref{tab:calibration}).

\begin{table}[H]
  \caption{Calibration sensitivity on Qwen3 14B$\to$32B / HellaSwag.
  Production values in bold.}
  \label{tab:calibration}
  \centering
  \small
  \setlength{\tabcolsep}{6pt}
  \begin{tabular}{llrr}
    \toprule
    Sweep & Setting & HellaSwag & $\Delta$ vs.\ prod.\ \\
    \midrule
    \multirow{5}{*}{$\lambda$ ($N{=}500$, FineWeb)}
      & 0       & 80.86 & $+0.13$ \\
      & 1e-4    & 80.88 & $+0.15$ \\
      & \textbf{0.01}    & \textbf{80.73} & --- \\
      & 0.1     & 79.75 & $-0.98$ \\
      & 1       & 64.94 & $-15.79$ \\
    \midrule
    \multirow{5}{*}{$N$ ($\lambda{=}0.01$, FineWeb)}
      & 50      & 79.09 & $-1.64$ \\
      & 100     & 79.72 & $-1.01$ \\
      & 200     & 80.44 & $-0.29$ \\
      & \textbf{500}     & \textbf{80.73} & --- \\
      & 1000    & 80.89 & $+0.16$ \\
    \midrule
    \multirow{3}{*}{Domain ($\lambda{=}0.01$, $N{=}500$)}
      & CodeAlpaca & 75.46 & $-5.24$ \\
      & Wikipedia  & 79.65 & $-1.05$ \\
      & \textbf{FineWeb-Edu}  & \textbf{80.70} & --- \\
    \bottomrule
  \end{tabular}
\end{table}

\paragraph{Calibration domain across benchmarks.}
We repeat the domain sweep across seven log-likelihood benchmarks,
evaluating each of the three domain-calibrated mappers on all of them
(Table~\ref{tab:calibration_grid}). Scoring every cell
by log-likelihood separates the calibration-domain effect from the additional
variance of generation decoding.

\begin{table}[H]
  \caption{Calibration domain $\times$ evaluation benchmark on Qwen3
  14B$\to$32B ($k{=}8$). Cells are accuracy, with retention against the
  FineWeb-Edu mapper on the same benchmark in parentheses.}
  \label{tab:calibration_grid}
  \centering
  \small
  \setlength{\tabcolsep}{5pt}
  \begin{tabular}{llrrr}
    \toprule
    Benchmark & Register & FineWeb-Edu & Wikipedia & CodeAlpaca \\
    \midrule
    HellaSwag     & commonsense           & 80.70 & 79.65 (98.7\%)  & 75.46 (93.5\%)  \\
    MMLU          & broad knowledge       & 78.07 & 78.05 (100.0\%) & 78.13 (100.1\%) \\
    ARC-Challenge & hard science QA       & 61.60 & 61.26 (99.5\%)  & 56.23 (91.3\%)  \\
    ARC-Easy      & easy science QA       & 83.63 & 83.08 (99.4\%)  & 80.43 (96.2\%)  \\
    PIQA          & physical commonsense  & 80.79 & 80.79 (100.0\%) & 79.49 (98.4\%)  \\
    BoolQ         & reading comprehension & 87.31 & 85.72 (98.2\%)  & 88.50 (101.4\%) \\
    WinoGrande    & coreference           & 68.98 & 69.22 (100.3\%) & 62.27 (90.3\%)  \\
    \midrule
    \emph{Mean}   &                       & 77.30 & 76.82 (99.4\%)  & 74.36 (95.9\%)  \\
    \bottomrule
  \end{tabular}
\end{table}

Retention stays above $90\%$ in every cell. The two out-of-domain corpora differ in
how their cost is spread. Wikipedia holds $99.4\%$ on average and varies
little from benchmark to benchmark. CodeAlpaca averages $95.9\%$ but ranges
over $11.1$\,pp, matching FineWeb-Edu on MMLU and BoolQ while costing the most
on coreference and hard science QA.

\section{Experimental setup details}
\label{app:setup}

This appendix collects the experimental-setup details referenced from
\S\ref{sec:setup} and \S\ref{sec:design}: the architectural axes of the
transfer pairs we study (Table~\ref{tab:families}), the benchmarks evaluated
(Table~\ref{tab:benchmarks}), benchmark coverage caveats, the perplexity
protocol used in our family tables, and per-pair mapper sizes
(Table~\ref{tab:mapper_sizes}).

\begin{table}[H]
  \caption{Transfer pairs studied with full architectural axes.
  All pairs are matched-KV ($n_{\text{kv}}^s = n_{\text{kv}}^t$ and
  $d_h^s = d_h^t$), the pre-condition for the linear-structure
  analysis in \S\ref{sec:linear_structure}, and use dense
  full-attention so every target layer receives mapped KV.}
  \label{tab:families}
  \centering
  \small
  \setlength{\tabcolsep}{3pt}
  \begin{tabular}{llcccc}
    \toprule
    Family & Source$\to$Target & Param ratio & KV heads (s$\to$t) & Head dim (s$\to$t) & Depth ratio \\
    \midrule
    Qwen3       & 8B, 14B $\to$ 32B    & 2.3--4$\times$ & 8$\to$8 (matched)  & 128$\to$128       & 1.6--1.8$\times$ \\
    Llama 3.1   & 8B $\to$ 70B         & 8.8$\times$    & 8$\to$8 (matched)  & 128$\to$128       & 2.5$\times$ \\
    Ministral 3 & 3B, 8B $\to$ 8B, 14B & 1.8--4.7$\times$ & 8$\to$8 (matched) & 128$\to$128       & 1.2--1.5$\times$ \\
    \bottomrule
  \end{tabular}
\end{table}

\paragraph{Benchmarks.}
We evaluate transfer quality on five accuracy benchmarks plus WikiText-2
perplexity, with CoQA for multi-turn handoff. Log-likelihood scoring follows the lm-evaluation-harness defaults.
MMLU uses 5-shot prompts and GSM8K uses 8-shot chain-of-thought prompts
\citep{wei2022cot}. The rest are zero-shot.

\begin{table}[H]
  \caption{Benchmarks evaluated for transfer quality. CoQA multi-turn
  (\S\ref{sec:multiturn}) is reported separately.}
  \label{tab:benchmarks}
  \centering
  \small
  \begin{tabular}{lll}
    \toprule
    Benchmark & Metric & Category \\
    \midrule
    ARC-Challenge \citep{clark2018arc}            & acc\_norm & Classification \\
    HellaSwag \citep{zellers2019hellaswag}        & acc\_norm & Classification \\
    WinoGrande \citep{sakaguchi2020winogrande}    & acc       & Classification \\
    MMLU 5-shot \citep{hendrycks2021mmlu}         & acc       & Classification \\
    GSM8K 8-shot \citep{cobbe2021gsm8k}           & acc       & Generation \\
    Perplexity (WikiText-2)                       & ppl       & Language modeling \\
    \bottomrule
  \end{tabular}
\end{table}

\paragraph{Coverage.}
All five accuracy benchmarks are evaluated on all six pairs at each pair's
selected $k$. Transfer perplexity is reported for the Qwen3 pairs and Llama 3.1.
Ministral~3 is reasoning-tuned, and its standalone WikiText-2 perplexity is
orders of magnitude above the other two families', so prefix-KV perplexity is
not a meaningful comparison for it and is not reported.
Missing entries appear as ``---'' in the per-family tables.

\paragraph{Perplexity protocol.}
Standard sliding-window perplexity does not isolate KV-cache quality from
the target model's own forward pass. We split WikiText-2 into
non-overlapping 2{,}048-token chunks and score each chunk's second 1{,}024
tokens conditioned on a 1{,}024-token prefix KV cache, taken from either
the target's own forward pass (standalone) or from the source via the
mapper (transfer). Both modes score identical continuation tokens,
isolating prefix-KV quality.

\paragraph{Mapper configuration.}
ARC-Challenge, HellaSwag and MMLU are evaluated at every $k$. WinoGrande,
the generation benchmarks and PPL are evaluated only at the selected $k$
and at $k{=}\text{all}$, to bound compute cost. For Ministral 3B$\to$8B, the criterion-optimal $k{=}20$ and
$k{=}\text{all}$ differ by $<0.6$\,pp on log-likelihood benchmarks. We
use $k{=}\text{all}$ to provide a single configuration across all benchmarks.

\begin{table}[H]
  \caption{Mapper size and storage by transfer pair, at the selected $k$
  used in \S\ref{sec:main}. Total parameter counts include both K and V
  heads across all target layers. ``all ($N$)'' means $k$ equals the
  number of source layers ($N$).}
  \label{tab:mapper_sizes}
  \centering
  \small
  \setlength{\tabcolsep}{6pt}
  \begin{tabular}{lcrr}
    \toprule
    Transfer & $k$ & Total params (K+V) & Storage \\
    \midrule
    Qwen3 8B$\to$32B       & 12        & 1.61\,B & 6\,GB   \\
    Qwen3 14B$\to$32B      & 8         & 1.07\,B & 4\,GB   \\
    Llama 3.1 8B$\to$70B   & 20        & 3.36\,B & 12\,GB  \\
    Ministral 3 3B$\to$8B  & all (26)  & 1.85\,B & 7\,GB   \\
    Ministral 3 8B$\to$14B & 12        & 1.01\,B & 4\,GB   \\
    Ministral 3 3B$\to$14B & 20        & 1.68\,B & 6\,GB   \\
    \bottomrule
  \end{tabular}
\end{table}

\paragraph{Serving cost.}
Mapper size is $2 \cdot L_t \cdot n_{\text{kv}}^t \cdot (k \cdot n_{\text{kv}}^s \cdot d_h^s) \cdot d_h^t$
for K and V together. It is set by target depth and head count and by $k$, and
does not grow with sequence length or cache size.

Mappers need not be GPU-resident: inference is one batched matmul per target
layer, so a serving stack can hold them on CPU memory or disk and page in the
active pair. At an assumed 25--50\,GB/s host-to-device link (PCIe Gen4/Gen5),
that is $\sim$80--480\,ms across the 4--12\,GB range, paid once when a pair
becomes active and amortized over its requests. These figures are computed
from size and bandwidth, not measured.

The ridge fit is directional, so each direction needs its own mapper and a
router over $P$ models covers up to $P(P{-}1)$ ordered pairs. At the average
size above ($\sim$6.5\,GB), 3-, 4- and 5-model fleets come to roughly 39, 79 and
131\,GB. The growth is quadratic in $P$, but the budget is disk or host memory,
not VRAM. The six mappers in the table span $3\times$ around that average.

\section{Run information}
\label{app:run_info}

This appendix records the calibration data, precision regime, MLP optimization
settings, and compute used. All experiments use a single $8\times$H100 node.

\begin{table}[H]
  \centering
  \small
  \begin{tabular}{lp{0.65\linewidth}}
    \toprule
    Parameter & Value \\
    \midrule
    Calibration data         & FineWeb-Edu, 500 sequences $\times$ 1{,}024 tokens, stride 4, $\sim$128K token-level observations per target head \\
    Precision                & bfloat16 (forward), float32 (covariance), float64 (analysis) \\
    Ridge regularization     & $\lambda = 0.01$ (production) \\
    Source feature           & $d_s = k\cdot n_{\text{kv}}^s\cdot d_h^s$ (cross-head concatenation) \\
    MLP baseline (\S\ref{sec:nonlin}) architecture & per (target layer, head, K$|$V): Linear($d_s$, 1024) $\to$ ReLU $\to$ Linear(1024, 1024) $\to$ ReLU $\to$ Linear(1024, $d_h^t$) \\
    MLP baseline optimization & Adam, lr $10^{-3}$, 20 epochs, MSE loss, batch size 4{,}096 \\
    Mapper fitting compute   & single $8\times$H100 node, $\sim$1 hour per pair end-to-end across matched-KV pairs (Qwen3 8B$\to$32B $\sim$47\,min, Qwen3 14B$\to$32B $\sim$52\,min, Llama 3.1 8B$\to$70B $\sim$50\,min, Ministral 3B$\to$8B $\sim$75\,min, 3B$\to$14B $\sim$84\,min, 8B$\to$14B $\sim$87\,min). Forming $\mathbf{X}^\top\mathbf{X}$ ($\mathcal{O}(N d_s^2)$) is computed once per target layer and shared across its heads \\
    Eval sharding            & HellaSwag/MMLU 16-way, ARC-Challenge 4-way \\
    \bottomrule
  \end{tabular}
\end{table}

\section{Extended main results}
\label{app:extended_results}

\begin{figure}[H]
  \centering
  \IfFileExists{figures/retention_bar.png}
    {\includegraphics[width=\linewidth]{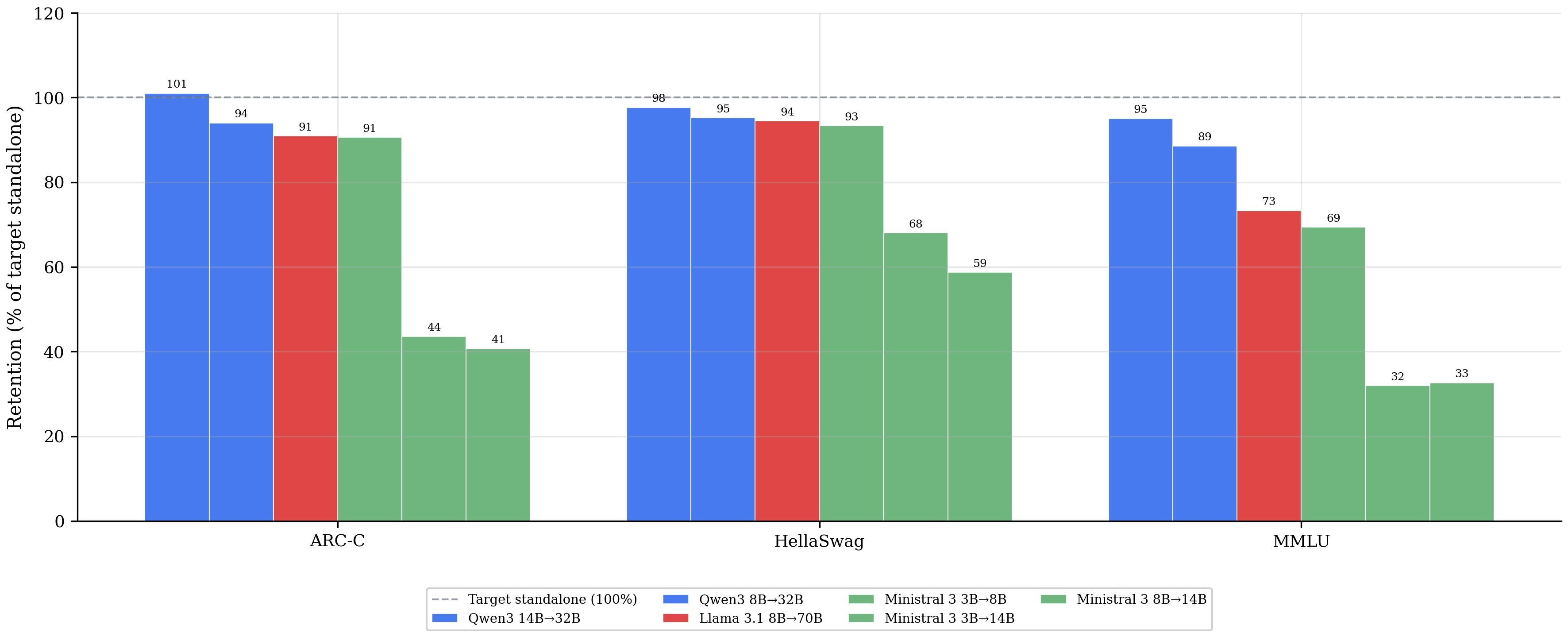}}
    {\fbox{\rule{\linewidth}{0pt}\rule{0pt}{4cm}}}
  \caption{Accuracy retention (\% of target standalone) across transfer pairs,
  on ARC-C, HellaSwag and MMLU. Dashed line: 100\% = target standalone.}
  \label{fig:retention}
\end{figure}

This appendix collects the per-family standalone vs.\ transfer
accuracies that back the headline numbers in \S\ref{sec:main}
(Table~\ref{tab:per_family}), the per-benchmark floor-normalized retention
(Table~\ref{tab:floor_normalized}), and a per-pair retention visualization
on ARC-C, HellaSwag and MMLU (Figure~\ref{fig:retention}).

\begin{table}[H]
  \caption{Per-family standalone and transfer results (full version of
  Table~\ref{tab:main_results}). Each block lists the target standalone
  baseline followed by the transfer accuracy from each source.}
  \label{tab:per_family}
  \centering
  \small
  \setlength{\tabcolsep}{4pt}
  \begin{tabular}{llrrrrrr}
    \toprule
    Family & Method & ARC-C & HellaSwag & WinoGrande & MMLU & GSM8K & PPL \\
    \midrule
    \multicolumn{8}{l}{\emph{Qwen3 (matched KV)}}\\
    & 32B standalone           & 61.01 & 82.65 & 70.01 & 82.17 & 95.15 & 6.79 \\
    & 14B standalone           & 60.49 & 78.87 & 69.30 & 78.86 & 95.00 & 7.63 \\
    & 8B standalone            & 56.83 & 75.00 & 64.88 & 75.42 & 91.66 & 8.67 \\
    & 14B$\to$32B (k=8)        & 61.60 & 80.70 & 68.98 & 78.09 & 90.98 & 7.33 \\
    & 8B$\to$32B (k=12)        & 57.34 & 78.67 & 63.69 & 72.74 & 65.50 & 7.98 \\
    \midrule
    \multicolumn{8}{l}{\emph{Llama 3.1 (matched KV)}}\\
    & 70B standalone           & 61.77 & 85.81 & 72.45 & 78.84 & 81.12 & 2.47 \\
    & 8B standalone            & 55.03 & 79.32 & 63.14 & 64.97 & 56.10 & 5.58 \\
    & 8B$\to$70B (k=20)        & 56.14 & 81.03 & 63.14 & 57.78 & 14.78 & 4.37 \\
    \midrule
    \multicolumn{8}{l}{\emph{Ministral 3 (matched KV)}}\\
    & 14B standalone           & 66.30 & 79.01 & 69.77 & 77.33 & 88.02 & --- \\
    & 8B standalone            & 59.81 & 76.92 & 67.01 & 72.97 & 84.38 & --- \\
    & 3B standalone            & 53.33 & 69.10 & 60.30 & 64.37 & 81.12 & --- \\
    & 3B$\to$8B (k=all)        & 54.18 & 71.78 & 61.17 & 50.61 & 30.86 & --- \\
    & 8B$\to$14B (k=12)        & 26.96 & 46.39 & 51.78 & 25.25 & 1.44  & --- \\
    & 3B$\to$14B (k=20)        & 28.92 & 53.75 & 51.62 & 24.73 & 2.81  & --- \\
    \bottomrule
  \end{tabular}
\end{table}

\paragraph{Floor-normalized retention.}
Table~\ref{tab:metric_anchors} reads the metric of \S\ref{sec:setup} at two
anchors on the Qwen3 32B target. The target's own ground-truth cache should
score 100\%. The 8B$\to$32B mapper stripped of RoPE separation and cross-layer
selection ($k{=}1$), the counterpart of the fourth row of
Table~\ref{tab:ablation}, scores at or below chance on ARC-C, WinoGrande and
MMLU, so a retention-like metric should read $\approx$0\% there.

\begin{table}[H]
  \caption{Metric anchors, as retention\,/\,floor-normalized retention on the
  Qwen3 32B target. The ablated mapper scores $48.5$ on WinoGrande against a
  chance floor of $50$, which retention credits as $69.2$\% and floor
  normalization places at $-7.7$\%.}
  \label{tab:metric_anchors}
  \centering
  \small
  \setlength{\tabcolsep}{4pt}
  \resizebox{\linewidth}{!}{%
  \begin{tabular}{lrrrrrr}
    \toprule
    Anchor & Avg & ARC-C & HellaSwag & WinoGrande & MMLU & GSM8K \\
    \midrule
    Ground-truth cache             & 99.9 / 99.9 & 100.0 / 100.0 & 100.0 / 100.1 & 100.0 / 100.0 & 100.0 / 100.0 &   99.3 / 99.3 \\
    $-$ RoPE $-$ cross-layer ($k{=}1$) & 35.1 / -0.5 &   37.8 / -5.5 &   38.4 / 11.7 &   69.2 / -7.7 &   28.9 / -2.1 &     1.1 / 1.1 \\
    \bottomrule
  \end{tabular}}
\end{table}

\begin{table}[H]
  \caption{Per-benchmark floor-normalized retention for the six matched-KV
  pairs. Chance floors are 25\% for ARC-C, HellaSwag and MMLU, 50\% for
  WinoGrande, and $\approx$0\% for GSM8K, so floor normalization leaves GSM8K
  unchanged.}
  \label{tab:floor_normalized}
  \centering
  \small
  \setlength{\tabcolsep}{5pt}
  \begin{tabular}{llrrrrrr}
    \toprule
    Family & Pair ($k$) & Avg & ARC-C & HellaSwag & WinoGrande & MMLU & GSM8K \\
    \midrule
    Qwen3       & 14B$\to$32B (8)              &   96.3 &  101.6 &   96.6 &   94.9 &   92.9 &   95.6 \\
    Qwen3       & 8B$\to$32B (12)              &   80.7 &   89.8 &   93.1 &   68.4 &   83.5 &   68.8 \\
    Llama 3.1   & 8B$\to$70B (20)              &   62.9 &   84.7 &   92.1 &   58.5 &   60.9 &   18.2 \\
    Ministral 3 & 3B$\to$8B (all)              &   65.9 &   83.8 &   90.1 &   65.7 &   53.4 &   36.6 \\
    Ministral 3 & 3B$\to$14B (20)              &   14.7 &    9.5 &   53.2 &    8.2 &   -0.5 &    3.2 \\
    Ministral 3 & 8B$\to$14B (12)              &   11.1 &    4.7 &   39.6 &    9.0 &    0.5 &    1.6 \\
    \bottomrule
  \end{tabular}
\end{table}

Normalization changes the per-benchmark ordering. WinoGrande has the highest
floor and moves the most, falling from 87.1\% to 58.5\% on Llama 3.1
8B$\to$70B. GSM8K is unchanged, its floor already being $\approx$0, and on the
strongest pair it now outranks WinoGrande.

\section{Prefill latency details}
\label{app:latency}

This appendix extends the single-pair measurement of \S\ref{sec:latency} to
seven pairs across the three families in both transfer directions.

\paragraph{Setup.}
Measured on a single $8\times$H100 node with NVLink, bf16, 50 warmup and 30
timed trials per cell. KV cache transfer latency includes the cross-GPU
transfers needed to move the source cache to the target. Re-prefill runs the
target's transformer body with \texttt{flash\_attention\_2}, excluding the LM
head. The mapper runs in eager mode, with no \texttt{torch.compile} or CUDA
graphs. At short sequences a
fixed floor from Python dispatch and cross-GPU transfers dominates the mapper's
wall time (14.0\,ms for the Qwen3 14B$\to$32B mapper).

\begin{table}[H]
  \caption{Re-prefill vs.\ KV cache transfer latency across seven pairs, each
  at its production $k$. Columns 3--4 are the 32{,}768-token point. The last
  column is the range over ten sequence lengths from 64 to 32{,}768.
  The mapper is faster in every one of the 70 cells.}
  \label{tab:latency_all}
  \centering
  \small
  \setlength{\tabcolsep}{5pt}
  \begin{tabular}{llrrr}
    \toprule
    \multicolumn{1}{c}{Family} & \multicolumn{1}{c}{Pair ($k$)} & \multicolumn{1}{c}{Transfer @32K} & \multicolumn{1}{c}{Re-prefill @32K} & \multicolumn{1}{c}{Speedup} \\
    \midrule
    \multicolumn{5}{l}{\emph{Small-to-large}} \\
    Qwen3       & 14B$\to$32B (8)    &    278\,ms &  6{,}975\,ms & 4.4--25.1$\times$ \\
    Qwen3       & 8B$\to$32B (12)    &    392\,ms &  6{,}975\,ms & 4.3--17.8$\times$ \\
    Llama 3.1   & 8B$\to$70B (20)    &    777\,ms & 11{,}562\,ms & 4.5--14.9$\times$ \\
    Ministral 3 & 3B$\to$14B (20)    &    396\,ms &  2{,}465\,ms & 3.8--6.2$\times$  \\
    Ministral 3 & 3B$\to$8B (all)    &    438\,ms &  1{,}764\,ms & 2.7--4.0$\times$  \\
    \midrule
    \multicolumn{5}{l}{\emph{Large-to-small}} \\
    Qwen3       & 32B$\to$14B (20)   &    427\,ms &  2{,}953\,ms & 3.3--6.9$\times$  \\
    Llama 3.1   & 70B$\to$8B (10)    &    216\,ms &  1{,}652\,ms & 2.8--7.6$\times$  \\
    \bottomrule
  \end{tabular}
\end{table}

\begin{figure}[H]
  \centering
  \IfFileExists{figures/latency_vs_seqlen.png}
    {\includegraphics[width=\linewidth]{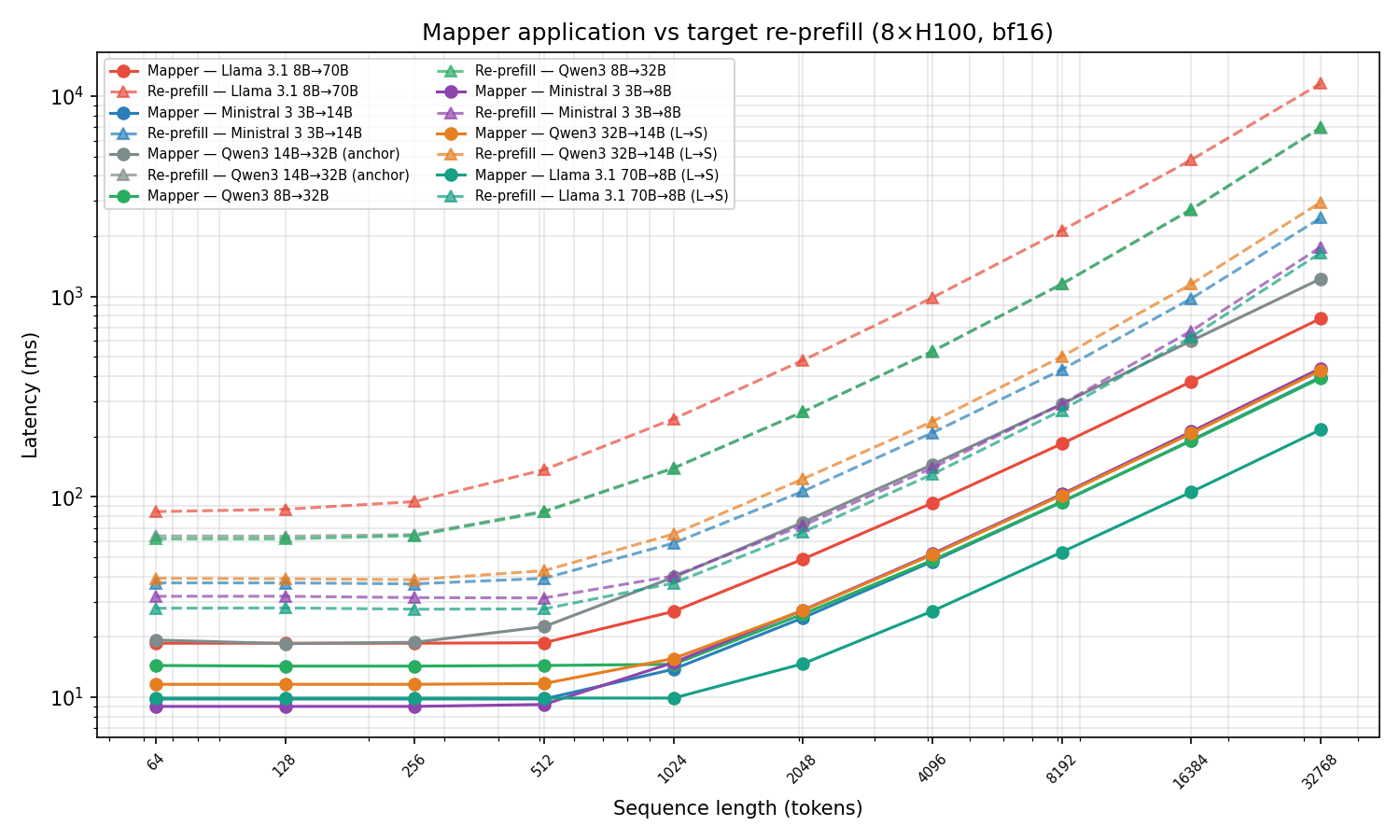}}
    {\fbox{\rule{\linewidth}{0pt}\rule{0pt}{4cm}}}
  \caption{Mapper application (solid) vs.\ target re-prefill (dashed) across
  sequence length, $8\times$H100 bf16. The mapper is near-flat at short
  sequences, where fixed dispatch cost dominates, then rises linearly.
  Re-prefill sits above the mapper for every pair at every sequence length.}
  \label{fig:latency}
\end{figure}

\paragraph{Caveats.}
Both conditions
operate on synthetic inputs, which isolates compute cost. End-to-end transfer
would also include shipping the mapped cache to the target process, which we
do not measure. Ministral 3 re-prefill is timed on the language-model
decoder body, excluding the vision tower, which is the correct comparison for
text cache transfer but not a full-model forward.

\section{Source-layer count selection}
\label{app:k_selection}

The source-layer count $k$ is chosen per pair on benchmarks that also appear
in the results. This appendix gives the selection protocol and bounds how much
a benchmark's own inclusion raises its reported accuracy.

\paragraph{Protocol.}
For each pair we sweep $k \in \{1,2,4,6,8,10,12,16,20,24,\text{all}\}$ and take
the $k$ maximizing the arithmetic mean of the log-likelihood benchmark
accuracies. The criterion nominally covers ARC-Challenge, HellaSwag,
WinoGrande and MMLU, but WinoGrande is absent at most $k$ for these six pairs,
so in practice the argmax is over ARC-Challenge, HellaSwag and MMLU. Near-ties
break toward larger $k$ for evaluation coverage: five pairs take the argmax
outright, and Ministral 3B$\to$8B takes $k{=}\text{all}$ over the argmax at
$k{=}20$, which it trails by $0.38$\,pp (Appendix~\ref{app:setup}).
GSM8K, CoQA and prefill
latency never enter selection: the argmax is computed over the three
benchmarks above, so they are holdouts by construction.

\paragraph{Leave-one-benchmark-out re-selection.}
We drop one selection benchmark, re-select $k$ by argmax on the remaining two,
and read the dropped benchmark at both the full-selection $k$ and the
leave-one-out $k$. This isolates how much a benchmark's own inclusion raises
its reported accuracy. The Ministral 3B$\to$8B rows use the data-driven
argmax ($k{=}20$), not its manual override to $k{=}\text{all}$.

\begin{table}[H]
  \caption{Leave-one-benchmark-out re-selection. $k$ is unchanged in 12 of 18
  folds. The held-out benchmark's accuracy moves by at most $2.49$\,pp
  (mean $0.30$\,pp), and always downward when it moves.}
  \label{tab:k_loo}
  \centering
  \small
  \setlength{\tabcolsep}{4pt}
  \begin{tabular}{llccr}
    \toprule
    Family & Pair & Full-select $k$ & LOO $k$ (drop HS / MMLU / ARC-C) & Max $\Delta$ (pp) \\
    \midrule
    Qwen3       & 8B$\to$32B  & 12 & 12 / 12 / 12 & $0.00$ \\
    Qwen3       & 14B$\to$32B & 8  & 4 / 10 / 6   & $0.61$ \\
    Llama 3.1   & 8B$\to$70B  & 20 & 20 / 24 / 20 & $1.45$ \\
    Ministral 3 & 3B$\to$8B   & 20 & 20 / 16 / 20 & $0.25$ \\
    Ministral 3 & 3B$\to$14B  & 20 & 20 / 20 / 20 & $0.00$ \\
    Ministral 3 & 8B$\to$14B  & 12 & 20 / 12 / 12 & $2.49$ \\
    \bottomrule
  \end{tabular}
\end{table}

The largest shift, $2.49$\,pp, occurs on Ministral 8B$\to$14B, a Tier 2 pair.
Across the four Tier 1 pairs it is at most $1.45$\,pp.

\paragraph{Held-out benchmarks.}
We evaluate each pair's selected mapper on three benchmarks that played no
part in selection (Table~\ref{tab:k_holdout}). Neither the fit nor $k$ is
changed, so this measures the reported configuration out of sample.

\begin{table}[H]
  \caption{Retention on three benchmarks never used to select $k$, at each
  pair's selected $k$. Retention is against the target's standalone accuracy
  on the same benchmark.}
  \label{tab:k_holdout}
  \centering
  \small
  \setlength{\tabcolsep}{8pt}
  \begin{tabular}{llrrrr}
    \toprule
    Family & Pair & PIQA & BoolQ & ARC-Easy & Mean \\
    \midrule
    Qwen3       & 14B$\to$32B            &  98.0\% & 101.9\% &  99.8\% & 99.9\% \\
    Qwen3       & 8B$\to$32B             &  97.1\% &  99.9\% &  94.4\% & 97.1\% \\
    Llama 3.1   & 8B$\to$70B             &  97.0\% &  96.2\% &  99.5\% & 97.5\% \\
    Ministral 3 & 3B$\to$8B              &  97.8\% &  95.1\% &  97.6\% & 96.8\% \\
    Ministral 3 & 3B$\to$14B             &  83.4\% &  57.4\% &  50.4\% & 63.7\% \\
    Ministral 3 & 8B$\to$14B             &  84.9\% &  44.4\% &  48.6\% & 59.3\% \\
    \bottomrule
  \end{tabular}
\end{table}

Retention is at or above in-sample retention for every pair, and the tier
structure of \S\ref{sec:main} reappears: the four Tier 1 pairs stay above
96\% and the two Tier 2 pairs below 64\%. PIQA and ARC-Easy are easier than the
selection benchmarks, so absolute levels are not comparable across the two
sets.

\end{document}